\documentclass[runningheads]{llncs}
\usepackage{graphicx}

\usepackage{tikz}
\usepackage{comment}
\usepackage{amsmath,amssymb} 
\usepackage{color}

\usepackage[accsupp]{axessibility}

\usepackage[width=122mm,left=12mm,paperwidth=146mm,height=193mm,top=12mm,paperheight=217mm]{geometry}

\usepackage[pagebackref,breaklinks,colorlinks]{hyperref}
\usepackage{booktabs}
\usepackage{enumitem}
\newcommand{\ra}[1]{\renewcommand{\arraystretch}{#1}}
\usepackage{todonotes}  
\usepackage[mathscr]{euscript}
 \let\mathscr\relax%
\usepackage[scr]{rsfso}

\begin{document}

\pagestyle{headings}
\mainmatter

\title{Motif Mining: Finding and Summarizing Remixed Image Content}

\titlerunning{Motif Mining}

\author{William Theisen\inst{1} \and
Daniel Gonzalez Cedre\inst{1} \and
Zachariah Carmichael\inst{1} \and
Daniel Moreira\inst{1} \and
Tim Weninger\inst{1} \and
Walter Scheirer\inst{1}
}
\authorrunning{W.~Theisen et al.}

\institute{University of Notre Dame, Notre Dame IN 46556, USA
\email{cse@nd.edu}\\
\url{http://cse.nd.edu}
}
\maketitle

\begin{abstract}
On the internet, images are no longer static; they have become dynamic content. Thanks to the availability of smartphones with cameras and easy-to-use editing software, images can be remixed (\textit{i.e.}, redacted, edited, and recombined with other content) on-the-fly and with a world-wide audience that can repeat the process. From digital art to memes, the evolution of images through time is now an important topic of study for digital humanists, social scientists, and media forensics specialists. However, because typical data sets in computer vision are composed of static content, the development of automated algorithms to analyze remixed content has been limited. In this paper, we introduce the idea of \emph{Motif Mining} --- the process of finding and summarizing remixed image content in large collections of unlabeled and unsorted data. In this paper, this idea is formalized and a reference implementation is introduced. Experiments are conducted on three meme-style data sets, including a newly collected set associated with the information war in the Russo-Ukrainian conflict. The proposed motif mining approach is able to identify related remixed content that, when compared to similar approaches, more closely aligns with the preferences and expectations of human observers.

\end{abstract}

\section{Introduction}

As the number of images posted online has grown, it has become impossible to summarize trends in this information by hand. Moreover, even though some computer vision algorithms have been proposed for this problem~\cite{yang2019deep,zhao2020deep}, the need for labelled ground truth presents a significant hurdle. The cost, in both money and time, is far too high. This is particularly evident when analyzing social trends, which often move so quickly that well-labelled data becomes obsolete by the time it is prepared. Moreover, the prevalence of remixed image content like memes, whereby an image is edited multiple times to remove information or incorporate other content, has raised questions about how related images should be associated. In this paper we describe the automated process of discovering trends in a large collection of remixed images as \textit{Motif Mining}. Several different communities are interested in this concept, including digital humanists studying new participatory art movements, computational social scientists studying the role of visual communication in conflicts, and media forensics specialists attempting to detect disinformation~\cite{shifman,ai4peace,reddit-data}.  

There are several challenges that must be overcome to achieve robust and accurate motif mining. To-date, the concept has been applied informally in the literature~\cite{zanne,beskow,dubey,icwsm}, leaving questions about optimization strategies that can be applied to the problem, as well as the structure of the output.  With respect to a viable algorithm, no image feature exists that works with both globally similar images and images that are similar only in small local regions --- what is commonly observed in remixed content. Additionally there has yet to be a large study of how the different combinations of image features and clustering algorithms affect the human perception of mined motifs. As the purpose of motif mining is to aid human observers, this is an important question to answer.

\begin{figure}[t!]
\centering
\includegraphics[width=1.0\textwidth]{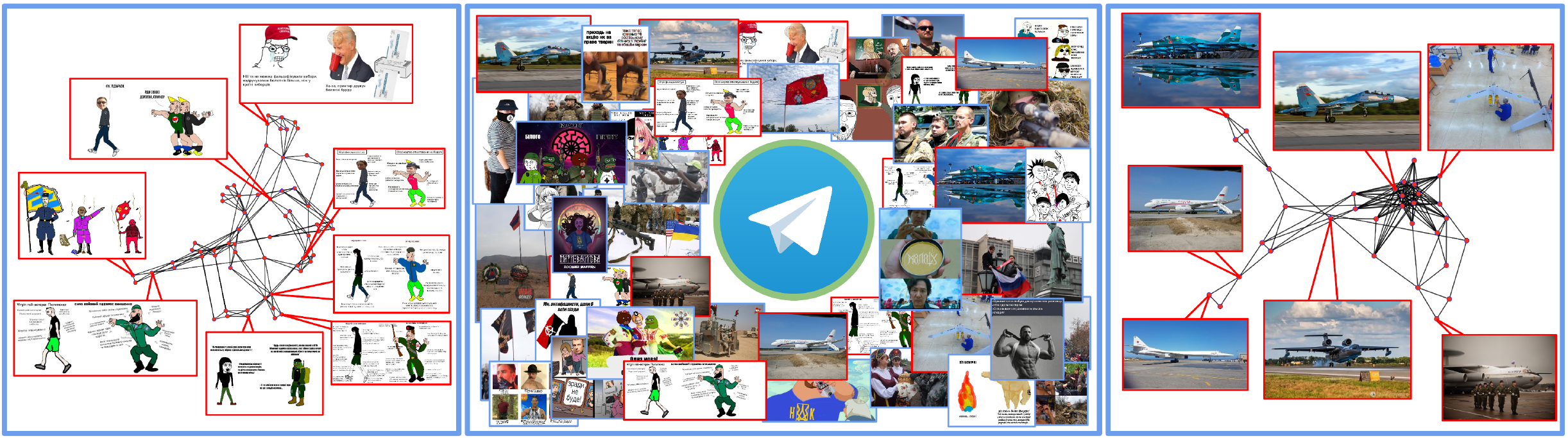}
\vspace{-6mm}
\caption{Given a large unsorted and unlabelled data set from social media, the motif mining strategy groups related content --- especially that which is remixed --- together in an unsupervised manner. Above are two motifs mined from a new data set of images related to the Russo-Ukrainian conflict collected from the Telegram platform~\cite{tele}.}
\vspace{-4mm}
\label{fig:tease}
\end{figure}

To address these challenges, this paper makes the following contributions:
\begin{enumerate}[noitemsep,nolistsep]
    \item A formal description of the problem of motif mining, providing a roadmap on how to more-easily discover salient trends in large, unsorted image corpora.
    \item A solution to this problem in the form of an end-to-end pipeline.\footnote{This system will be open-sourced pending publication of this work.}
    \item An image feature strategy for this problem combining both local and global information to aid in human-preferred clustering.
    \item A new data set of over half of a million posts and remixed and static images collected from Telegram over the past six years, including the beginning of the 2022 invasion of Ukraine by the Russian Federation.
    \item An empirical study of the proposed pipeline, including comparisons to related approaches on three data sets containing remixed and static image content. 
\end{enumerate}  

\noindent\textbf{Related Work.}
Within computer vision, motif mining is most closely related to content-based image retrieval (CBIR)~\cite{dubey2021decade}, with differences in their respective inputs and purposes.
CBIR takes as input (i) an image of interest (namely, a query) and (ii) a corpus of potentially related images (namely, the gallery), and aims to retrieve the images from the gallery that are similar to the query, sorting them from the most to the least similar, according to a well-defined
similarity criteria.
Depending upon the CBIR system user’s intent, the similarity criteria may range from retrieving images that are semantically similar (\textit{e.g.}, images that depict the same type of objects as the query), to retrieving images that are near-duplicates (\textit{e.g.}, images that are minor variations of the query, thus sharing portions of pixels that come from the same imaging pipeline).

Regardless of the intent, CBIR solutions focus on reducing the semantic gap between the values of the image pixels and the user’s objective.
To overcome this gap, typical CBIR solutions operate in a multi-level image representation approach.
At the lower level, either local or global features are extracted from the pixel values.
While local features describe portions of the image that depict interesting visual phenomena (such as corners, edges, blobs, etc.), global features individually describe the entire image content.
As expected, while many local features are usually extracted to represent a single image, only one to a few global features are used to represent the same content.

Independent of being local or global, features are either handcrafted (\textit{i.e.}, carefully engineered by a CBIR expert who targeted a particular set of visual phenomena), or learned (\textit{i.e.}, they are the outcome of a data-driven machine learning solution).
Popular examples of handcrafted local features include SIFT~\cite{lowe2004distinctive} and SURF~\cite{bay2008speeded}, while more recent  learned local feature approaches include LIFT~\cite{yi2016lift}, DELF~\cite{noh2017large}, and LISRD~\cite{pautrat2020online}.
They are most commonly used for 
performing image registration, when different pictures of the same object or scene taken from distinct standpoints are stitched together in a post-processing operation.
Handcrafted global features, in turn, comprise the concatenation of patch-wise LBP~\cite{ojala2002multiresolution} and PHASH~\cite{monga2006perceptual}, only to name a few.
Their common use case is also the retrieval of near-duplicate images, or different captures of the same object.
More recently, the use of intermediary convolutional layers (before the fully connected layers) of popular architectures of neural networks as global image descriptors has become possible, including features from VGG~\cite{vgg}, ResNet~\cite{he2016deep}, and MobileNet~\cite{mobile} (hereafter MOBILE).
These features are useful for the retrieval of semantically similar images because they leverage the learned content classification ability of their respective networks.

One level up, CBIR solutions aim to index the low-level features to reduce their inverted file index (IVF) storage space through feature compression and by speeding up the retrieval of feature-wise k-nearest neighbors.
The standard solution in feature indexing is based on the retrieval of approximate nearest neighbor (ANN) features for each one the query's features, supported  by optimized product quantization (OPQ~\cite{opq}) of all the features.
FAISS~\cite{faiss} is a popular library that implements different indexing strategies, including OPQ.
Finally, at the highest level, once a set of features from the gallery images is retrieved for all the query's features, voting schemes based on IVF are used to find the gallery images that are the most similar to the query.
Leveraging the voting count, gallery images can be sorted from most to least similar, constituting the desired output: a ranked list of images similar to the query.

In contrast to CBIR, motif mining takes as input a large corpus of images of interest only; there is no query to take as a reference.
Moreover, rather than returning a ranked list of similar images, the purpose of motif mining is to find different motifs, \textit{i.e.}, groups (or clusters) of images whose similarity of interest is not known at execution time.
This aspect is something to be discovered, as part of the problem formulation, being sometimes semantic (in the case of conceptually similar images) and sometimes based on pixel values (in the case of images sharing templates, such as memes containing stock character macros~\cite{shifman}), or even both (such as the example motif depicted on the left-hand side of Fig.~\ref{fig:tease}).

Despite their differences in both input and purpose, we show that motif mining borrows from CBIR the combined use of local and global low-level features (because there is no clear definition of whether semantically similar images or near-duplicates are desired), as well as the best feature indexing strategies.
In a similar fashion, recent work such as
the algorithm described by Niu et al.~\cite{niu2018multi} have focused on merging features of different modalities, such as visual and textual content.
We instead combine only visual features.
At the upper level, in turn, image clustering is performed, similar to the methods proposed for the specific case of image-based memes by Zannettou et al.~\cite{zanne}, Beskow et al.~\cite{beskow}, Dubey et al.~\cite{dubey}, and Theisen et al.~\cite{icwsm}.
The novelty of the present work includes on a new formalization of the problem with reference to human-acceptable output to guide a new algorithmic design. Further, our approach is not constrained to just remixed content like memes, and it works as a method to identify visually similar static images in an unsupervised manner as well. 

\section{Formalization of Motif Mining}

Traditionally, computer vision problems have been framed as an optimization over some metric calculated in reference to ground truth data for a task. Although this provides high-quality baselines for comparison, there is often little effort expended on demonstrating that higher metric scores actually result in more useful output for human observers for tasks like image retrieval. As an alternative, the methods and procedures of visual psychophysics from psychology have been recently recommended as a way to use human behavioral responses to evaluate algorithms~\cite{webster2018psyphy,RichardWebster_2018_ECCV}. Taking insight from these prior works, we can formalize motif mining.

\noindent\textbf{The Motif Mining Problem.} The purpose of motif mining is to allow human observers to quickly gain insights about visual trends in a large collection of unsorted and unlabeled data. The most common method for finding multiple trends in a given data set is clustering. However, as the purpose of motif mining is to aid people, the clusters must be optimized around some feedback mechanism. We differentiate between clusters produced with no human feedback and human-preferred clusters, and we structure our experiments around this idea.

For an example of a useful cluster,  the right-hand side of Fig.~\ref{fig:tease} shows a number of different airplanes, several of which are fighter jets. This example is drawn from the current Russo-Ukrainian conflict. An increase in the number of militaristic images being posted online might prefigure action in a conflict~\cite{ai4peace} and could also potentially leak useful and/or damaging intelligence to third parties. The Ukrainian government recently addressed this concern specifically, with ``Ukraine's defense minister, Oleksii Reznikov, [...] calling on viewers to share images of Russia's assault'' and ``a local Telegram channel urged its 400,000 subscribers to `carefully film' and share video of passing Russian troops so Ukrainian fighters could hunt them down''~\cite{ukr-social}. These examples were taken from 851 motifs mined from a subset of 16,433 images from the Ukrainian data set collected from Telegram~\cite{tele}. The ideal number of clusters and the distribution of the images across them is best formalized as a clustering optimization problem with the task accuracy being derived from human feedback.

\noindent\textbf{Optimization for Human Observers.}
Given a large, unlabelled corpus of images, our goal is to automatically discover trends in this data set by classifying those images that can be thought of as being ``conceptually similar'' or ``derived from the same picture'' in some intuitive sense.
Because our classification task is both inherently intuitive and difficult to formally specify, and because the quantity of data far exceeds any human annotators' ability to manually label, we develop an unsupervised system for clustering these images together and verify them \emph{a posteriori} by humans.

Here, our formal framework specifies our data set of images as a weighted graph $\mathcal{G} = \left (\mathcal{V}, \mathcal{E}, w: \mathcal{E} \to \mathbb{R}_+ \right )$,
where $\mathcal{V}$ is a vertex set for a graph $\mathcal{G}$ and $\mathcal{E}$ is its edge set. Here, $w: \mathcal{E} \to \mathbb{R}_+$ denotes a function that assigns positive, real-valued weights to each of the edges of $\mathcal{G}$.
The vertices in this graph represent images from the data set and whose weighted edges represent the strength of the similarity between two adjacent images.
Within this framework, the task becomes computing an unsupervised clustering of $\mathcal{V}$ such that it disagrees as little as possible with what human observers expect.
We test this using the \textit{intruder detection task}~\cite{weninger2012document}. This means finding some subset $\mathcal{C}$ of the power set of the vertices
$\mathcal{P}(\mathcal{V})$ such that, for a given pair of clusters $c_1, c_2 \in \mathcal{C}$, if a human were presented with $k$ images from $c_1$ and one intruder-image from $c_2$, the human would be able to pick the intruder.
We define this formally as follows:
\begin{equation}
\begin{aligned}\label{eq:problem-formulation}
\min_{\mathcal{C} \in \mathfrak{P}(\mathcal{V})}
&{\left ( \sum_{c \neq \tilde{c} \in \mathcal{C}}~\sum_{v_1, \dots v_k \in c}~\sum_{\tilde{v} \in \tilde{c}}
\left (1 - H(v_1, \dots v_k, \tilde{v})\right )\gamma(c, \tilde{c}) \right )},\\
\end{aligned}
\end{equation}
\noindent where (i)~$k \in \mathbb{N}_+$, (ii)~$\mathcal{G} = (\mathcal{V}, \mathcal{E}, w: \mathcal{E} \to \mathbb{R}_+)$ is the graph, (iii)~$\mathfrak{P}(\mathcal{V})$ is the set of partitions of $\mathcal{V}$, (iv)~$\mathcal{C}$ is a clustering on $\mathcal{G}$, (v)~${v_1, \dots v_k}$ all belong to the same cluster $c$, (vi)~$\tilde{v}$ belongs to a different cluster $\tilde{c}$, (vii)~$H: \mathcal{V}^{k+1} \to \{0, 1\}$ returns $1$ iff $\tilde{v}$ is correctly identified by a human, and (viii)~$\gamma(\cdot, \cdot)$ is a normalizing factor.

\begin{figure}[t]
\centering
\includegraphics[width=0.98\textwidth]{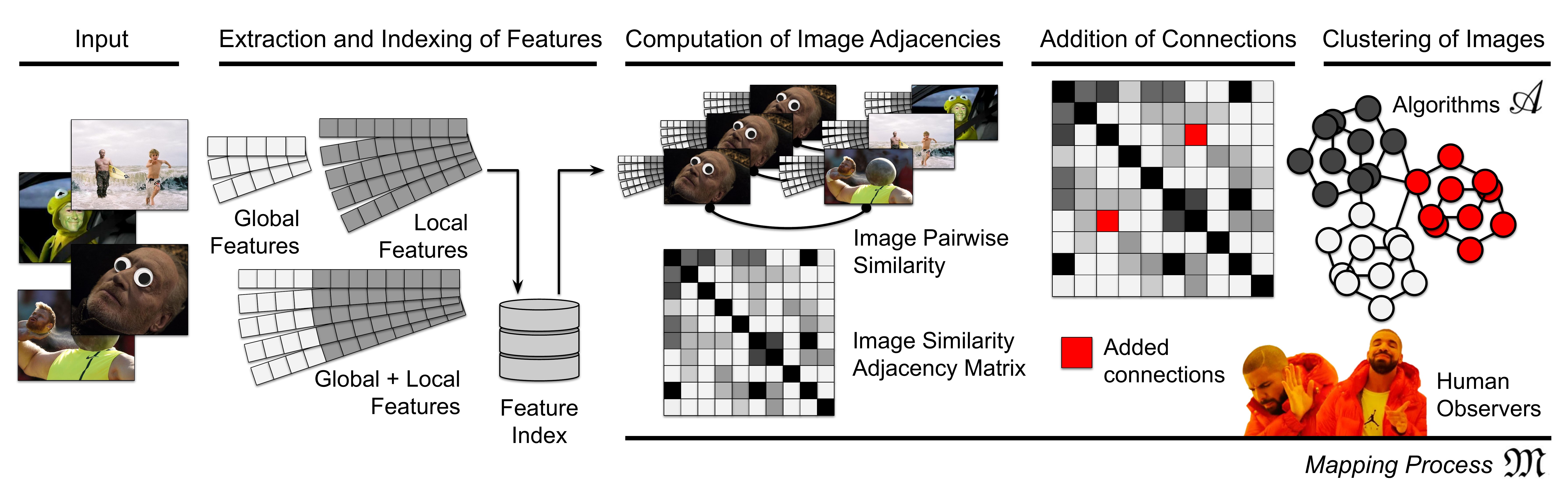}
\vspace{-4mm}
\caption{Motif mining pipeline.
The process starts with a large set of images of interest as input and ends with the mined motifs, \textit{i.e.}, clusters of the images that summarize remixed content.
Four steps constitute the pipeline: extraction and indexing of low-level image features, computation of image adjacencies according to their pairwise similarities, computation of additional similarity connections, and clustering of the images.
Human observers provide feedback on the quality of the motifs.
}
\vspace{-4mm}
\label{fig:pipeline}
\end{figure}

In order to specify this graph $\mathcal{G}$, a mapping process $\mathfrak{M}$, with some corresponding parameters, is needed to map the image corpus onto a weighted graph.
With this in mind, the problem can be further thought of as an optimization task like Eq.~(\ref{eq:problem-formulation}) for each weighted graph realizing the given data set.
Thus, Eq.~\ref{eq:problem-formulation} will be minimized for a given clustering $\mathcal{C}$ of $\mathcal{G}$ only when human observers agree with the quality of the clustering.
In order to find such a high-quality clustering, it is computationally infeasible to simply enumerate all the possible clusterings.
Instead, a more principled approach would parameterize a clustering algorithm $\mathscr{A}$ suited for clustering weighted graphs and perform this optimization task over the set of $\mathscr{A}$'s parameters.
This would, however, require an enormous number of human observers to check the quality of each clustering produced during this optimization.

In this paper, we employ a small variety of effective graph clustering algorithms and check their performance directly against the human observers for a few different realizations of the graph produced by $\mathfrak{M}$.
Every time $\mathfrak{M}$ produces a weighted graph, we apply one of the clustering algorithms $\mathscr{A}_i$ to the graph and evaluate the quality of those clusters.
This heuristic approach is a step in the direction of finding the true minimum alluded to in Eq.~\ref{eq:problem-formulation}.

\section{Implementation of Motif Mining}
\label{sec: pipeline}
Fig.~\ref{fig:pipeline} summarizes our pipeline for motif mining.
Each one of the steps depicted within it are detailed as follows.

\noindent\textbf{Extraction and Indexing of Features.}
The first step towards motif mining is to determine the kind of features that will be used to generate the vectors in the index.
This decision is informed primarily by the types of motifs an observer wishes to find in the data set.
Global and local feature extraction methods will produce intuitively different results in the types of images returned by a given query and therefore the types of motifs mined.
PHASH~\cite{phash} features, for instance, will return only duplicate or near-duplicate images.
MOBILE~\cite{mobile} and VGG~\cite{vgg} features will return images that are similar globally in a semantic sense; this often manifests in something akin to images that all contain airplanes, though those airplanes may be different shapes, sizes, positions, and styles.
SURF~\cite{bay2008speeded} features lead to connections that may be 
visually distinct and share only some very small local information, such as a hand gesture or a logo on a flag.
 
It is the combination of these two types of features that creates compelling and robust connections.
Quite frequently, SURF features will not return near duplicate images in the top of their query results as there are smaller, more subtle matches somewhere in the image.
 However, global visual similarity is very apparent and useful to human observers.
 If, for example, military groups begin to post edited pictures of tanks or inflammatory extremist symbols, human observers will want to quickly identify these trends.
 To capture both of these cases, each image has a single global feature extracted in addition to its SURF features.
 The full set of global features is then subjected to Principal Component Analysis (PCA)~\cite{pearson1901liii} and each vector is reduced to $16$ dimensions, to match the length of the smallest of the descriptors, namely PHASH.
 A further discussion about this parameter is provided in Section~\ref{app:cent_tag} of the Appendix.

 This global ``tag'' is then appended to each of the respective images' SURF features.
 Therefore, global context for any given image is incorporated into all of its individual local feature vectors.
 The effect of adding this global tag can be seen in both the airplane motif on the far right of Fig.~\ref{fig:tease}, and the flag motif in the third row and third column of Fig.~\ref{fig:full_tri}.

 The local features come to the fore when the motif is a smaller object in an image, such as the Indonesian ballot boxes with ``KPU'' written on them as seen in column three, row two of Fig.~\ref{fig:full_tri}.

To index the features,
the proposed pipeline is implemented using the FAISS \cite{faiss} library as a foundation.
Available within FAISS, OPQ~\cite{opq} allows for efficient mass-vector indexing and retrieval.
Similar to the work previously done by Theisen et al. \cite{icwsm}, an 
IVF is made with 256 centroids (an exploration of how the number of centroids affects the index and resulting graphs can be found in Section~\ref{app:cent_tag} of the appendix).
This index, once built, provides the function used to generate the graph that is then clustered. Readers familiar with the workings of 
OPQ may at this point wonder why the centroid clusters that are inherent to OPQ and already generated
by FAISS are not just used for the end product, without the extra hassle of producing 
another graph on top of the index.
This is discussed in a subsection of Section 4.
Building this index allows us to quickly construct an approximate affinity matrix, and a subsequent graph, by leveraging the efficiency of OPQ.

\noindent\textbf{Computation of Image Adjacencies.}
We need to perform image pairwise comparisons to compute image similarities prior to establishing the motif clusters.
To compute similarities between images leveraging the index built in the previous step, we need to select a set of images as starting points to query their features from the index.
Given IVF indices are built at the feature level, the indexed features have a many-to-one relationship with their respective source images.
As a consequence, retrieved features need to be ``mapped'' back to the image from which they were extracted, since we are interested in image-level similarity.
Considering the querying of the features of a selected starting-point image, each result $r \in \mathcal{R}$ is a tuple $(f, i, d)$, where $f$ is a feature corresponding to an image $i$, and $d$ is the distance between $f$ and the queried feature as computed by OPQ.
If we focus on the subset $\mathcal{R}_i$ of the retrieved features that belong to image $i$, we can compute the similarity $s_i$ between that image and the selected starting-point one as follows:

\begin{equation}\label{eq:ranking_sim_score}
    s_i = \sum_{(f, i, d) \in \mathcal{R}_i} 1 - \tanh{(d)}.
\end{equation}

\noindent We elect to use the nonlinear operator $\tanh{(\cdot)}$ as it is nicely bounded within the interval $[0, 1)$ for all non-negative $d$.

Intuitively, the nonlinear weighting rewards smaller distances and penalizes distances more harshly as they become larger.

Note that this similarity computation is a loop only for the local features (including the tagged ones).
For the global features, since their relationship is 1:1 with the source image, we can simply take the single distance value returned by OPQ and compute the similarity score in a similar fashion to Eq.~\ref{eq:ranking_sim_score}.

To realize a graph out of image similarity computations, we create an $N \times N$ adjacency matrix, whose each row/column is determined by one of the $N$ images in the data set.

We thus define entry $(i, j)$ of this matrix to be the similarity value computed through Eq.~\ref{eq:ranking_sim_score}.
The entries in this matrix will then correspond to weighted edges between vertices, which represent the images.
Several strategies have been explored for selecting the starting-point images and filling in the scores in the matrix.
Prior work~\cite{icwsm} has simply taken a smaller subset of the images in the set and hoped that the resulting connections were diverse enough to form a representative graph.
In this work, we instead continue selecting a random subset of isolated images in the graph (\textit{i.e.}, images whose columns and rows within the adjacency matrix sum up to zero), until all the images are visited.
This method is cheaper than querying all $N$ images while still eliminating any isolated vertices, unlike prior work.
Note that this does not ensure one singular connected component, though some features (\textit{e.g.}, SURF) do still lead to fully-connected graphs on smaller data sets.

\noindent\textbf{Addition of Image Connections.}
The fact that our image similarity graph is not fully connected imposes an \emph{a priori} constraint on any proposed vertex clusters since most clustering algorithms treat the presence of multiple components as a strong prior.
So, if the connected components do not capture the images' underlying homophily (\textit{i.e.}, if perceptually-similar images are distributed across different connected components), then this will act as a strong prior biasing even the best algorithms away from a high-quality clustering.
To combat this effect, one could connect the disjoint components according to some principled (\textit{i.e.}, data-driven) or heuristic schema.
Here, we propose a random baseline connection schema, specified by an Erd\H{o}s-R\'{e}nyi~\cite{erdos1960evolution} model over the connected components, and the \emph{Best} and \emph{Average} heuristic approaches.

The baseline Erd\H{o}s-R\'{e}nyi model takes a parameter $p$, which specifies the probability of adding an edge between any two components.
The weights assigned to these new edges are proportional to the average weights of the edges in the two components being connected.

We find $p$ so that the expected number of new edges added to the graph is linear in the initial number of components.
This avoids needlessly changing the density of the graph.

The \emph{Best} and \emph{Average} connection strategies work similarly to the Erd\H{o}s-R\'{e}nyi approach but with different strategies for determining when components get connected with each other.
Given $N_C$ total components and a proposed pair of components $C_i$ and $C_j$,
these algorithms compare the components by extracting their vertices' associated feature vectors.
The cosine similarity of these feature vectors then determines the ``similarity'' between the two components.
The \emph{Best} approach assigns a similarity to the pair $(C_i, C_j)$ based on the \emph{most similar} pair of vertices found from $C_i$ and $C_j$.
The \emph{Average} approach, by contrast,
assigns a similarity to $(C_i, C_j)$ based on the \emph{average similarity} of their corresponding vertex pairs.

In either case, we then find a threshold $\theta$ so that the number of pairs $(C_i, C_j)$ with similarity scores above $\theta$ is proportional to $N_C$.
Those pairs of components are then connected as follows:
the \emph{Best} adds edges between those vertices that had the most similar feature vectors; the \emph{Average} approach randomly connects $k$-many pairs of vertices (by default, $k = 1$).

These new edges are weighted in proportion to the components' similarity.

\noindent\textbf{Clustering of Images.} A variety of unsupervised clustering techniques, well-suited to finding communities in weighted graphs, were tested. Louvain clustering, Markov clustering, and Spectral clustering. Spectral clustering was chosen so that results can be compared to the prior literature~\cite{icwsm}.

The \emph{Louvain} method for community detection maximizes the modularity of the graph --- a measure comparing the density within and across clusters --- using a two-stage iterative optimization.

\emph{Markov} clustering, on the other hand, is a random-walk based clustering algorithm that computes transition probabilities between the vertices of a weighted graph by modeling random walks over the graph as Markov chains.

Finally, \emph{Spectral} clustering refers to a very popular approach to clustering data according to the eigenvalues of some similarity matrix computed over the data.
In the context of clustering for graphs, the Spectral approach involves applying $k$-means clustering to the vertices of the graph using the $k$ largest eigenvalues of the graph's Laplacian matrix $L = D - A$ as features, where $A$ is the graph's (weighted) adjacency matrix, and $D$ is its diagonal.

These clustering algorithms provide an unsupervised way of exposing underlying trends in the data --- the way in which remixed and static images would naturally be agglomerated by a human.
These clusters are what human observers will judge during evaluation, so choosing a high-quality algorithm with sensible parameters should noticeably impact our performance.
While it is not obvious which of these algorithms should be most similar to how a human would cluster the data, we can intuit that Louvain, with its modularity optimization, should maximize the separation between clusters and thus ensure as much difference between images from different clusters as is possible to infer from the graph.
Markov clustering, which relies on the idea of distributing flow among the edges of the graph, might produce a clustering with comparatively ``softer'' boundaries.
Finally, eigenvalue-based approaches like Spectral clustering, known to be related to probabilistic diffusion processes on graphs~\cite{nadler2005nips}, rely on finding cuts in the graph based on its Laplacian's eigenvalues, and thus may intuitively fill a gap in-between Louvain and Markov clustering.
Which of these, if any, most closely corresponds to human intuition can be revealed only through human evaluation.

\section{Experiments and Results}

In total, 252 different configurations of the Motif Mining pipeline described in Sec.~\ref{sec: pipeline} were tested in order to identify the most effective ones.
 Figure~\ref{fig:acc_comps} provides a summary of all of the these results, while Section~\ref{app:acc_scores} of the appendix presents them individually in a more detailed tabular form. Of primary interest is a particular combination's accuracy on the Imposter-Host test, which serves as a proxy as to whether or not the produced motifs are valuable to human observers. To describe the configurations we use the format \textit{feature\_type-connection\_type-clustering\_type} where feature type is one of: PHASH, MOBILE, VGG, SURF, SURF\_PHASH, SURF\_MOBILE, SURF\_VGG. Connection type is one of: average (avg), best, Erd\H{o}s-R\'{e}nyi (er), or unconnected (reg). Finally the clustering method is one of Louvain, Markov, or Spectral.

\begin{figure}[!t]
\centering
\includegraphics[width=1.0\textwidth]{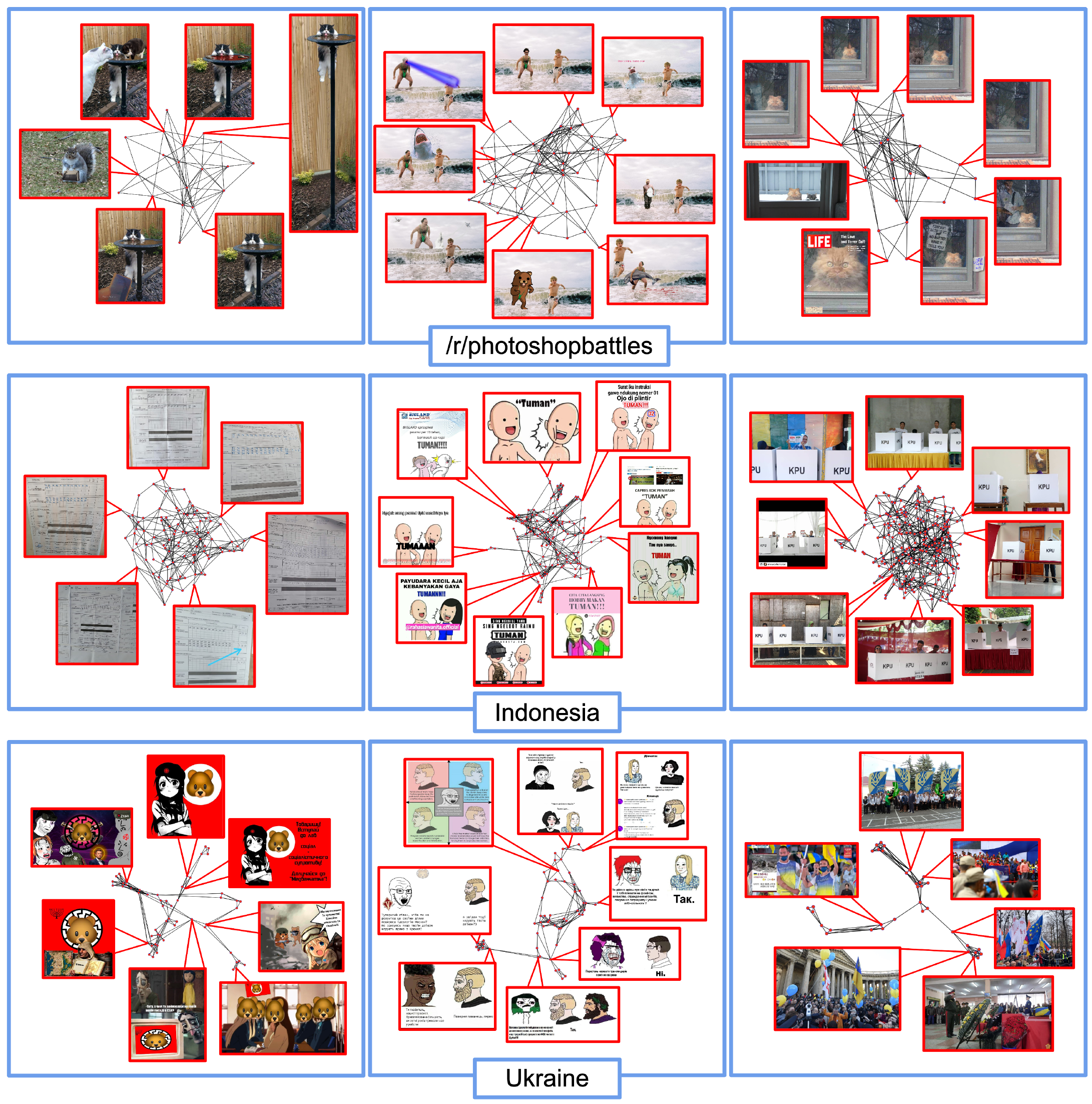}
\caption{Nine motifs discovered from the Reddit Photoshop Battles dataset~\cite{reddit-data}, 2019 Indonesian National Election dataset~\cite{icwsm}, and a newly collected dataset associated with the Russo-Ukrainian conflict. A variety of global motifs, local motifs, and remixed content can be seen across the nine examples. From cats to alleged voting fraud, the new pipeline can discover a diverse array of motifs in any data set.}
\vspace{-4mm}
\label{fig:full_tri}
\end{figure}

\begin{figure}[t]
\centering
\includegraphics[width=1.0\textwidth]{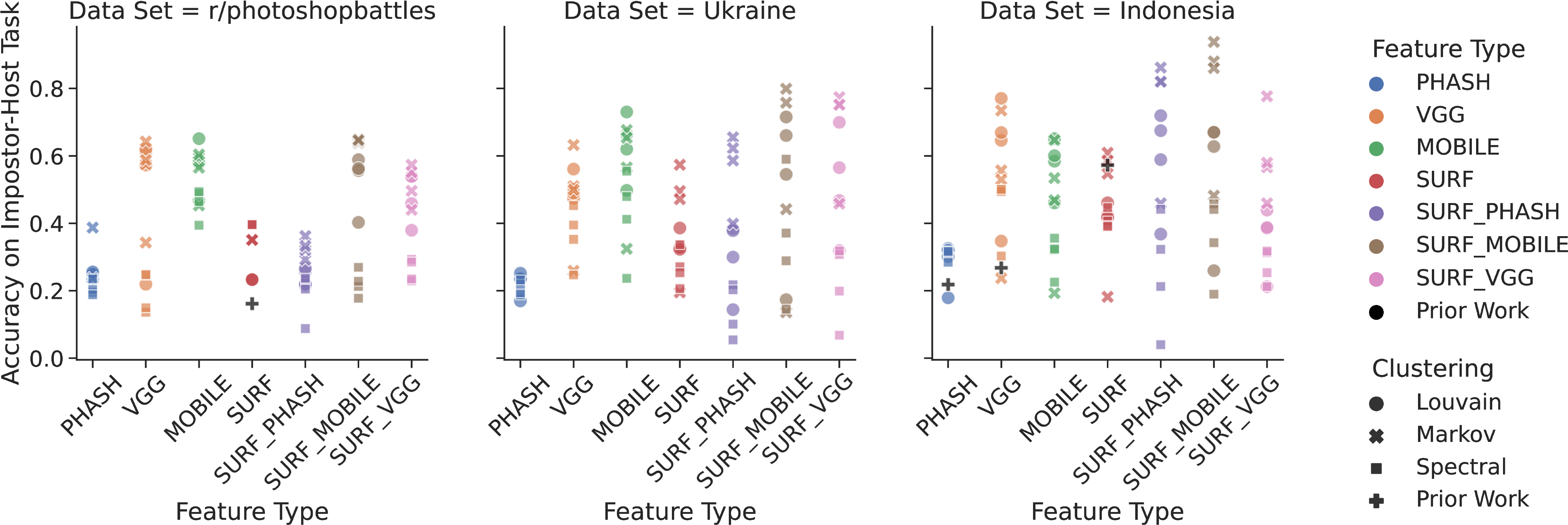}
\vspace{-4mm}
\caption{The accuracy scores of the Imposter-Host test across the three data sets. Each of the three clustering methods is noted with a different shape. For further distinction with respect to the connection algorithm used we point to Section~\ref{app:acc_scores} of the Appendix.}
\vspace{-4mm}
\label{fig:acc_comps}
\end{figure}

\noindent\textbf{Data Sets.} The first benchmark data set used for experiments was the \textit{Reddit Photoshop Battles}~\cite{reddit-data} data set, an image remix-specific set for content-based image retrieval and image clustering. It consists of 10,586 images taken from threads on Reddit's r/photoshopbattles subreddit. It proves to be a particularly challenging collection due to the diversity of the submitted images as seen in Fig.~\ref{fig:full_tri}. This data set provides the `purest' collection of remixed images, as the purpose of the subreddit is to take a piece of a `donor' image and insert it into a variety of different sub-images. The second set consisted of 44,612 images (including memes) related to the 2019 Indonesian Presidential election (shortened to `Indonesia'), as used in \cite{icwsm}, to allow for an additional point of comparison to previous work. 

A new data set of interest to the human rights community was also collected (referred to as `Ukraine'). Scraped from Telegram and starting in the year 2016 and continuing through the beginning of March 2022, it focuses on content related to the Russo-Ukrainian conflict. Containing controversial images relating to nationalism, fascism, xenophobia, racism, homophobia, and the growth of militia and para-military groups, it provides an unparalleled look into the growth of online tensions surrounding the conflict. Comprised of 665,725 images and 721,441 posts, it is relevant to use for the testing of tools aimed at aiding human rights activists in the fight against online hate and disinformation. Out of this a subset of 16,433 images was selected as a test set for the purpose of these experiments.

With the help of experts within Ukraine,
a list of Telegram users was compiled. The channels associated with these users were then scraped,
thus forming the data set. A list of these users may be found in Section~\ref{app:users} of the Appendix. Both the data set and the scraping tools will be released alongside the publication of this work. Each post consists of a JSON file containing, in addition to the post title and text, relevant meta-data such as the post date, view count at the time of scraping, image links associated with the post (Telegram allows more than one image per post), and the raw image files.

\noindent\textbf{Assessing Cluster Relatedness with an Imposter-Host Test.} As the goal of the pipeline is to create clusters for humans to review, testing the ``accuracy'' of the clusters requires human input. To this end, we use a version of the \emph{Imposter-Host} test~\cite{imp-host} outlined in Theisen et al. \cite{icwsm}, a standard way of evaluating classification tasks in a human-centric way. This test consisted of asking 50 Amazon Mechanical Turk~\cite{turk} workers to find which image out of a set of five was the most different 25 times (with 5 of those 25 being control questions). Four of the images shown were from a single ``host'' cluster, and the fifth image --- that the Turk workers are tasked with identifying --- was taken from a randomly selected ``imposter'' cluster. Intuitively, the more related images in any given cluster are, the easier it is for the Turk workers to pick out the imposter image. 
Since selections are made from a set of five images, the baseline accuracy (computed by randomly picking one of the five images) is 20\%. This was done for each of the 84 different possible combinations of feature types, connection types, and clustering methods and was done for all 3 data sets.

Reddit, the smallest data set, seems to imply that a global feature yields the best results in terms of observer accuracy per Figure~\ref{fig:acc_comps}. However as explained in Section 4.1, the global features are constrained by the number of centroids that the index is initialized with. For a smaller data set like Reddit, containing only 10,588 images, 256 clusters is enough to achieve high accuracy scores (this intuition is based on there being 186 reddit threads comprising the data set, which if we take as a proxy for classes implies that 186 motifs mined would be the perfect answer). However as the data sets grow larger the available number of clusters stays at 256. This is in contrast to the globally tagged features, which continue to grow in the number of clusters (though not strictly in proportion). In this work, increasing accuracy scores are seen amongst the global-local features as the number of images increases. Should the number of images continue to increase it seems intuitive to believe that the accuracy of the global features would begin to decrease as more and more visually diverse images will have to be fit to a maximum of 256 clusters, while the global-local features allow for more clusters as the data grows.

\begin{table}[t]
\small
\centering
\ra{1.3}
\begin{tabular}{@{}lcccccccccc@{}}
\toprule
                                     &&& \cite{icwsm} (Top) &&& Ours (Top - Spectral) &&& Ours (Top) \\ \midrule
\multicolumn{1}{l}{Reddit - PHASH} &&& N/A              &&& 23.53\%                                    &&& \textbf{38.73\%}                   \\ 
\multicolumn{1}{l}{Reddit - VGG}   &&& N/A              &&& 24.79\%                                    &&& \textbf{64.25\%}                   \\ 
\multicolumn{1}{l}{Reddit - SURF}  &&& 16.15\%          &&& \textbf{39.62\%}                              &&& \textbf{39.62\%}                   \\ 
\multicolumn{1}{l}{Indonesia - PHASH}   &&& 21.83\%          &&& 31.81\%                                    &&& \textbf{32.53\%}                   \\
\multicolumn{1}{l}{Indonesia - VGG}     &&& 26.79\%          &&& 50.07\%                                    &&& \textbf{77.05\%}                   \\
\multicolumn{1}{l}{Indonesia - SURF}    &&& 57.25\%          &&& 44.66\%                                    &&& \textbf{60.94\%}                   \\ \bottomrule
\end{tabular}
\caption{\label{tab:icwsm_comp}The top accuracy scores on the Imposter-Host task compared against Theisen et al.~\cite{icwsm}. Shown are the top scores from the proposed pipeline using the Spectral clustering method to allow for a fair comparison to how the clustered graph is generated in previous work, followed by the top score for any combination. This shows that, except for the Reddit - SURF combination, Spectral clustering maximize accuracy.}
\vspace{-4mm}
\end{table}

With respect to the Imposter-Host task accuracy scores, state of the art results are achieved. As seen in Table~\ref{tab:icwsm_comp} prior literature on the Reddit data set achieved only a 16.15\% accuracy, worse than random chance, claiming that ``due to the visual complexity of
the data set, they [Turk Workers] were able to find connections that weren’t
intended to link images''~\cite{icwsm}. The new pipeline achieves a highest accuracy score of 65.11\% using the \texttt{mobile-best-louvain} combination. This represents an increase of nearly 50 percentage points (48.96). In addition to top results on the Reddit data set, double-digit improvements were also seen on the Indonesian data. Prior work found an accuracy of 57.25\% on the data. The new pipeline achieves an accuracy of 93.81\% with \texttt{surf\_mobile-best-markov} (a difference of 36.56 percentage points).

On the new Ukrainian/Russian data set, the accuracy scores are similarly high. A top score of 79.91\% is seen using the \texttt{surf\_mobile-er-markov} method. Fig.~\ref{fig:acc_comps} further shows the tagged features' high accuracy as the data set grows.

There is, however, an important caveat to this result. Much as Theisen et al.~\cite{icwsm} found that Spectral clustering produced a singular ``mega-cluster'' that contained the vast majority of the images and thus making the method undesirable, the Markov clustering method had a similar issue, inadvertently skewing the results towards the higher side.
Instead of placing all of the odd-out images into a single massive cluster, it placed each into its own individual cluster.
With the Imposter-Host test requiring at least 4 images in a host cluster, these individual image clusters were ignored.
This left a number of clusters containing only near duplicates which would thus improve the Imposter-Host accuracy scores.
Due to this quirk, unless observers were interested in small near-duplicate clustering, we would recommend using the Louvain clustering method in a real-world implementation of motif mining as it results in a much more even and usable distribution of images with still state-of-the-art accuracy scores on the Imposter-Host task (a plot showing this distribution may be seen in Section~\ref{app:img_distr} of the Appendix).

\noindent\textbf{Underlying Graph and Cluster Structures.}
Although the new graph creation method ensures that there are no isolated vertices in the graph, it is not guaranteed to produce a single connected component. Instead, the graph contains several interesting patterns emerging from the type of feature used in the creation of the index, and from choices relating to the initialization of the index.

To explore this further an experiment was run in which each index was recreated on the Reddit data set with a different number of centroids (128, 256, 512, 1024).
The global feature types always resulted in graphs with a number of components equal to the number of centroids the index was initialized with. The global features resulting in connected components equal to the number of centroids implies that all the querying step of the pipeline is doing is exposing the underlying centroid-space that FAISS has already prepared and therefore could be done away with entirely (see Section~\ref{app:cent_tag} in the Appendix for details).

Although SURF features by themselves resulted in a highly connected graph and global features resulted in simply mirroring the underlying cluster space FAISS had already computed, the ``tagged'' (\textit{i.e.}, global + local) features resulted in a higher number of components than the number of centroids, implying that further sub-structures of similar images were discovered within the centroids. The tagged features producing more components percolates down the pipeline to the clustering step, where the clusters using these features give a better image-cluster spread.

\noindent\textbf{Qualitative Results.}
During the processing of the Ukrainian data an initial test of the pipeline was run with a \texttt{surf\_mobile-reg-louvain} configuration, which we recommend as the best option when running in the wild due to the combination of speed, accuracy, and image distribution. The results were interesting enough that we proceeded to run it on all three data sets producing the results seen in Fig.~\ref{fig:tease} and Fig.~\ref{fig:full_tri}. In Fig.~\ref{fig:full_tri} the leftmost cluster in the top row shows five images of a cat drinking out of a birdbath and a spurious match on a squirrel with a briefcase.  The middle motif is of a man chasing a child out of the ocean. Here we see much stronger examples of remixing, with laser beams, sharks, and a bear all being added. The final motif is of a cat sitting at a door. This grouping highlights the usefulness of the local matching as one of the images has a stark global contrast from all of the others but the local features allow the matching on the shared cat's head.

The second row illustrates three motifs from the Indonesian data set. Again we can see a strong cluster of remixed content in the middle. On the left can be seen images of voting tallies, which the Prabowo campaign used as alleged evidence of fraud in a failed attempt to contest the 2019 election \cite{prabowo}. The third cluster again demonstrates why local features are extremely useful in motif mining. Many different images of Indonesians at ballot boxes were present in this motif but only shared the locally similar 'KPU' logo on the boxes.

Finally the bottom row of Fig.~\ref{fig:full_tri} shows results from the Ukrainian dataset. A cartoon bear head used as a logo by one of the extremist meme channels is found in the left-most panel. Note that this imagery is often superimposed over a Sonnenrad, which is a co-opted Nazi rune \cite{black-sun}. In the middle are several variations of a Ukrainian version of the Yes-Chad meme \cite{yes-chad}. On the right is a cluster of Ukrainian coloured flags. Of particular interest is the photo-shopped image top-center, showing school children bearing a number of flags.

\section{Conclusions}

The newly proposed pipeline, combined with a novel combination of image features, achieves state-of-the-art results on the motif mining problem. With increases reaching nearly 50 percentage points improvement over previous methods, it demonstrates a path forward for aiding human responses to emerging trends in online social media. In addition, a new data set has been collected from Telegram to allow further bench-marking in this space. Its timely release will allow researchers to gain unparalleled views into the increasing tensions online between Ukrainian and Russian actors, mirroring the growing tensions happening on the ground.

\section{Acknowledgements}
This material is based on research sponsored by the Defense Advanced Research Projects Agency (DARPA) and Air Force Research Laboratory (AFRL) under agreement numbers FA8750-16-2-0173 and FA8750-20-2-1004 and supported by USAID Cooperative Agreement \#7200AA18CA00059. 

\newpage

\clearpage
\bibliographystyle{splncs04}
\bibliography{egbib}

\begin{thebibliography}{10}
\providecommand{\url}[1]{\texttt{#1}}
\providecommand{\urlprefix}{URL }
\providecommand{\doi}[1]{https://doi.org/#1}

\bibitem{turk}
Amazon mechanical turk. \url{https://www.mturk.com/}

\bibitem{bay2008speeded}
Bay, H., Ess, A., Tuytelaars, T., Van~Gool, L.: Speeded-up robust features
  ({SURF}). Elsevier Computer Vision and Image Understanding  \textbf{110}(3),
  346--359 (2008)

\bibitem{beskow}
Beskow, D., Kumar, S., Carley, K.M.: The evolution of political memes:
  Detecting and characterizing internet memes with multi-modal deep learning.
  Information Processing and Management  \textbf{57}(2) (2020)

\bibitem{dubey}
Dubey, A., Moro, E., Cebrian, M., Rahwan, I.: Memesequencer: Sparse matching
  for embedding image macros. In: In Proceedings of WWW'18 (2018)

\bibitem{dubey2021decade}
Dubey, S.R.: A decade survey of content based image retrieval using deep
  learning. IEEE Transactions on Circuits and Systems for Video Technology
  (2021)

\bibitem{erdos1960evolution}
Erd{\H{o}}s, P., R{\'e}nyi, A., et~al.: On the evolution of random graphs.
  Publ. Math. Inst. Hung. Acad. Sci  \textbf{5}(1),  17--60 (1960)

\bibitem{opq}
Ge, T., He, K., Ke, Q., Sun, J.: Optimized product quantization for approximate
  nearest neighbor search. In: 2013 IEEE Conference on Computer Vision and
  Pattern Recognition. pp. 2946--2953 (2013). \doi{10.1109/CVPR.2013.379}

\bibitem{ukr-social}
Harwell, D., Lerman, R.: How ukrainians have used social media to humiliate the
  russians and rally the world.
  \url{https://www.washingtonpost.com/technology/2022/03/01/social-media-ukraine-russia/}
  (2022)

\bibitem{he2016deep}
He, K., Zhang, X., Ren, S., Sun, J.: Deep residual learning for image
  recognition. In: Proceedings of the IEEE conference on computer vision and
  pattern recognition. pp. 770--778 (2016)

\bibitem{mobile}
Howard, A., Sandler, M., Chu, G., Chen, L., Chen, B., Tan, M., Wang, W., Zhu,
  Y., Pang, R., Vasudevan, V., Le, Q.V., Adam, H.: Searching for mobilenetv3.
  In: In Proceedings of ICCV'19 (2019)

\bibitem{faiss}
Johnson, J., Douze, M., J{\'e}gou, H.: Billion-scale similarity search with
  {GPUs}. IEEE Transactions on Big Data  \textbf{7}(3),  535--547 (2019)

\bibitem{phash}
Klinger, E., Starkweather, D.: phash: The open source perceptual hash library.
  \url{https://www.phash.org} (2013)

\bibitem{lowe2004distinctive}
Lowe, D.: Distinctive image features from scale-invariant keypoints. Springer
  International Journal of Computer Vision  \textbf{60}(2),  91--110 (2004)

\bibitem{monga2006perceptual}
Monga, V., Evans, B.L.: Perceptual image hashing via feature points:
  performance evaluation and tradeoffs. IEEE Transactions on Image Processing
  \textbf{15}(11),  3452--3465 (2006)

\bibitem{reddit-data}
Moreira, D., Bharati, A., Brogan, J., Pinto, A., Parowski, M., Bowyer, K.,
  Flynn, P., Rocha, A., Scheirer, W.: Image provenance analysis at scale. IEEE
  Transactions on Image Processing  \textbf{27},  6109--6123 (08 2018).
  \doi{10.1109/TIP.2018.2865674}

\bibitem{nadler2005nips}
Nadler, B., Lafon, S., Kevrekidis, I., Coifman, R.: Diffusion maps, spectral
  clustering and eigenfunctions of fokker-planck operators. In: Weiss, Y.,
  Sch\"{o}lkopf, B., Platt, J. (eds.) Advances in Neural Information Processing
  Systems. vol.~18. MIT Press (2005),
  \url{https://proceedings.neurips.cc/paper/2005/file/2a0f97f81755e2878b264adf39cba68e-Paper.pdf}

\bibitem{niu2018multi}
Niu, Y., Lu, Z., Wen, J.R., Xiang, T., Chang, S.F.: Multi-modal multi-scale
  deep learning for large-scale image annotation. IEEE Transactions on Image
  Processing  \textbf{28}(4),  1720--1731 (2018)

\bibitem{noh2017large}
Noh, H., Araujo, A., Sim, J., Weyand, T., Han, B.: Large-scale image retrieval
  with attentive deep local features. In: IEEE International Conference on
  Computer Vision. pp. 3456--3465 (2017)

\bibitem{ojala2002multiresolution}
Ojala, T., Pietikainen, M., Maenpaa, T.: Multiresolution gray-scale and
  rotation invariant texture classification with local binary patterns. IEEE
  Transactions on Pattern Analysis and Machine Intelligence  \textbf{24}(7),
  971--987 (2002)

\bibitem{prabowo}
Paddock, R.C.: Indonesia court rejects presidential candidate’s voting fraud
  claims.
  \url{https://www.nytimes.com/2019/06/27/world/asia/indonesia-widodo-prabowo-election-fraud.html}
  (2019)

\bibitem{pautrat2020online}
Pautrat, R., Larsson, V., Oswald, M.R., Pollefeys, M.: Online invariance
  selection for local feature descriptors. In: Springer European Conference on
  Computer Vision. pp. 707--724 (2020)

\bibitem{pearson1901liii}
Pearson, K.: Liii. on lines and planes of closest fit to systems of points in
  space. The London, Edinburgh, and Dublin philosophical magazine and journal
  of science  \textbf{2}(11),  559--572 (1901)

\bibitem{webster2018psyphy}
RichardWebster, B., Anthony, S., Scheirer, W.: Psyphy: A psychophysics driven
  evaluation framework for visual recognition. IEEE Transactions on Pattern
  Analysis and Machine Intelligence  \textbf{41}(9),  2280--2286 (2018)

\bibitem{RichardWebster_2018_ECCV}
RichardWebster, B., Kwon, S.Y., Clarizio, C., Anthony, S.E., Scheirer, W.J.:
  Visual psychophysics for making face recognition algorithms more explainable.
  In: Proceedings of the European Conference on Computer Vision (ECCV)
  (September 2018)

\bibitem{shifman}
Shifman, L.: Memes in Digital Culture. The MIT Press (2013)

\bibitem{vgg}
Simonyan, K., Zisserman, A.: Very deep convolutional networks for large-scale
  image recognition. In: International Conference on Learning Representations
  (2015)

\bibitem{black-sun}
Sonnenrad.
  \url{https://www.adl.org/education/references/hate-symbols/sonnenrad}

\bibitem{tele}
{Telegram FZ LLC and Telegram Messenger Inc.}: Telegram.
  \url{https://telegram.org/}

\bibitem{icwsm}
Theisen, W., Brogan, J., Thomas, P.B., Moreira, D., Phoa, P., Weninger, T.,
  Scheirer, W.: Automatic discovery of political meme genres with diverse
  appearances. Fifteenth International AAAI Conference on Web and Social Media
  \textbf{15},  714--726 (2021)

\bibitem{imp-host}
Weninger, T., Bisk, Y., Han, J.: Document-topic hierarchies from document
  graphs. In: In Proceedings of CIKM'12 (2012)

\bibitem{weninger2012document}
Weninger, T., Bisk, Y., Han, J.: Document-topic hierarchies from document
  graphs. In: Proceedings of the 21st ACM international conference on
  Information and knowledge management. pp. 635--644 (2012)

\bibitem{yang2019deep}
Yang, X., Deng, C., Zheng, F., Yan, J., Liu, W.: Deep spectral clustering using
  dual autoencoder network. In: IEEE/CVF Conference on Computer Vision and
  Pattern Recognition. pp. 4066--4075 (2019)

\bibitem{ai4peace}
Yankoski, M., Theisen, W., Verdeja, E., Scheirer, W.J.: Artificial intelligence
  for peace: An early warning system for mass violence. Towards an
  International Political Economy of Artificial Intelligence pp. 147--175
  (2021)

\bibitem{yes-chad}
Yes chad. \url{https://knowyourmeme.com/memes/yes-chad}

\bibitem{yi2016lift}
Yi, K.M., Trulls, E., Lepetit, V., Fua, P.: {LIFT}: Learned invariant feature
  transform. In: Springer European conference on computer vision. pp. 467--483
  (2016)

\bibitem{zanne}
Zanneettou, S., Caulfield, T., Blackburn, J., Cristofaro, E.D., Sirivianos, M.,
  Stringhini, G., Suarez-Tangil, G.: On the origins of memes by means of fringe
  web communities. In: In Proceedings of IMC'18 (2018)

\bibitem{zhao2020deep}
Zhao, J., Lu, D., Ma, K., Zhang, Y., Zheng, Y.: Deep image clustering with
  category-style representation. In: Springer European Conference on Computer
  Vision. pp. 54--70 (2020)

\end{thebibliography}

\clearpage
\appendix

{\centering\textbf{\huge Appendices}
\vspace{0.25in}}

\section{Runtimes}\label{app:run}

The various run times for the pipeline vary widely depending on what feature type is used for extraction. PHASH and SURF features were the quickest due to their ease of parallelization and in SURF's case, its ability to run on a GPU. MOBILE features are noticeably slower but still much faster than VGG features, which took more than twice as long as MOBILE features on the Indonesia data set. It is for this reason primarily the we recommend against using VGG features. PHASH and SURF features, while fast, achieved low scores on both the Reddit and Ukraine data set in their individual forms, and slightly higher in the ``tagged'' combination of the two. Surprisingly the SURF\_PHASH score on the Indonesia data set was quite high and comparable to the top scores. It's unclear whether this was a fluke, due to some quirk in the data set, or due to the increasing size of the data set. More work needs to be done to explore this but if speed was of the absolute essence it would be worth trying this feature combination to explore a sufficiently large data set. If speed does not matter as much we recommend a variety of the MOBILE features. It is important to keep in mind that the MOBILE features by themselves will be stuck to a number of clusters equal to the number of centroids the OPQ index is initialized with and thus we prefer the SURF\_MOBILE combination which allows for more clusters and thus achieving a better image/cluster ratio.

Adding images to the index is extremely quick and should not be a serious consideration when exploring motif mining. On the other hand, the graph creation, connection, and clustering has serious run time implications. Local feature querying is significantly slower than the global features due to the voting required to map back to the images from the features retrieved from the index. On top of this, the graph connection process is costly. Do note that the run-times for this portion of the pipeline include all three connection methods and in practice only one would need to be used. Even with the connection methods sped up using dynamic programming the BEST and AVG connection methods still averaged seven hours approximately for the Indonesia data set. We don't believe this is worth the CPU time due to no noticeable increase in the accuracy scores on the resulting graphs. Human observers seem not to notice whether or not a graph has been connected prior to clustering. The clustering run times include all three methods on all four graphs and therefore, in an implementation in which one were to run only a single combinations, are of no serious concern.

An important note is that these run-times are a sample size of 1 and therefore should only be used as rough guidelines to how long one might expect the pipeline to run, but times might vary depending on the hardware and other activities on the machine. These run times were collected on a machine with an Intel Xeon E5-2620 v3 (12) @ 3.200GHz (CPU), 256 GB of RAM and a Titan X and Titan Z (GPUs). 

\begin{table}[h]
\tiny
\centering
\begin{tabular}{|c|c|c|c|}
\hline
Feature Extraction (Total Runtime) & Reddit (10586)      & Ukraine (16433)     & Indonesia (44612)   \\ \hline
PHASH (CPU, PE=6)                  & 00:01:02 (00:01:17) & 00:00:17 (00:00:36) & 00:01:59 (00:02:47) \\ \hline
MOBILE (CPU, PE=1)                 & 00:41:10 (00:41:38) & 00:39:51 (00:40:27) & 02:04:54 (02:06:27) \\ \hline
VGG (GPU, PE=1)                    & 01:41:31 (01:41:54) & 02:25:53 (02:26:27) & 06:58:22 (06:59:48) \\ \hline
SURF (GPU, PE=6)                   & 00:05:08 (00:20:27) & 00:06:20 (00:25:14) & 00:14:05 (01:02:55) \\ \hline
SURF\_PHASH (GPU/CPU, PE=6)        & 00:06:30 (00:23:36) & 00:06:31 (00:29:12) & 00:14:59 (01:15:10) \\ \hline
SURF\_MOBILE (GPU/CPU PE=1)        & 01:14:49 (1:39:40)  & 00:56:30 (01:19:26) & 02:46:13 (03:47:54) \\ \hline
SURF\_VGG (GPU, PE=1)               & 02:17:33 (02:30:37) & 02:35:49 (02:58:35) & 07:33:56 (08:35:24) \\ \hline
\end{tabular}
\caption{\label{tab:runtime1}CPU, GPU indicates which device the feature extraction was performed on. PE gives the number of parallel processes used during the feature extraction. Due to its low overhead, PHASH is trivial to parallelize which decreases the time needed to extract features. Times are expressed in the ``hours:minutes:seconds'' format.}
\end{table}

\begin{table}[h]
\centering
\begin{tabular}{|c|c|c|c|}
\hline
Index Add    & Reddit (10586) & Ukraine (16433) & Indonesia (44612) \\ \hline
PHASH        & 00:00:02       & 00:00:02        & 00:00:02          \\ \hline
MOBILE       & 00:00:02       & 00:00:03        & 00:00:03          \\ \hline
VGG          & 00:00:03       & 00:00:02        & 00:00:02          \\ \hline
SURF         & 00:00:31       & 00:00:42        & 00:02:46          \\ \hline
SURF\_PHASH  & 00:00:31       & 00:00:43        & 00:01:55          \\ \hline
SURF\_MOBILE & 00:00:32       & 00:00:44        & 00:02:43          \\ \hline
SURF\_VGG    & 00:00:25       & 00:00:43        & 00:02:08          \\ \hline
\end{tabular}
\caption{\label{tab:runtime2}
The time spent to add all feature vectors to the index, for each feature type.
Times are expressed in the ``hours:minutes:seconds'' format.}
\end{table}

\begin{table}[h]
\tiny
\centering
\begin{tabular}{|c|c|c|c|}
\hline
Graph/Cluster Creation & Reddit (10586)             & Ukraine (16433)            & Indonesia (44612)          \\ \hline
PHASH                  & 00:00:04/00:12:05/00:02:45 & 00:00:04/00:14:29/00:02:35 & 00:00:13/02:14:17/00:16:17 \\ \hline
MOBILE                 & 00:00:05/00:17:24/00:03:47 & 00:00:05/00:25:55/00:03:23 & 00:00:11/02:30:59/00:06:08 \\ \hline
VGG                    & 00:00:04/00:12:08/00:03:13 & 00:00:05/00:28:13/00:03:45 & 00:00:09/02:22:41/00:13:45 \\ \hline
SURF                   & 00:43:46/00:00:28/00:07:29 & 00:49:41/00:04:05/00:08:34 & 04:51:25/00:39:42/00:21:42 \\ \hline
SURF\_PHASH            & 00:32:10/00:17:18/00:03:28 & 00:59:48/01:44:55/00:05:30 & 07:11:42/14:02:41/00:13:53 \\ \hline
SURF\_MOBILE           & 00:43:46/00:13:07/00:03:59 & 01:08:49/02:33:48/00:05:38 & 07:57:32/14:53:36/00:15:54 \\ \hline
SURF\_VGG              & 00:30:07/00:09:58/00:01:53 & 01:04:06/01:50:55/00:09:00 & 05:43:41/18:56:54/00:13:42 \\ \hline
\end{tabular}
\caption{\label{tab:runtime3}
Times spent to create the clusters and mine the motifs, for each feature type.
Times are expressed in the ``hours:minutes:seconds'' format.}
\end{table}

\section{Imposter-Host Accuracy Tables}\label{app:acc_scores}

Below are the full tabular results for the Imposter-Host test accuracy scores. The scores marked as N/A were invalid due to there be a number of clusters equal to the number of images in the data set and therefore no purpose in running the task. There is no apparent pattern in which graph connection method observers preferred and for that reason we mostly recommend against using them, for run-time purposes. However, if time is of no concern a number of top scores were produced using the BEST connection method and could be tried. While Markov clustering produced the highest scores we recommend the Louvain method due to the healthier spread of images amongst the clusters.

\begin{table}[h]
\centering
\begin{tabular}{|c|c|c|c|}
\hline
Reddit       & Louvain                                                                                               & \multicolumn{1}{c|}{Markov}                                                                           & \multicolumn{1}{c|}{Spectral}                                                                         \\ \hline
PHASH        & \begin{tabular}[c]{@{}c@{}}23.46\% - AVG\\ 25.18\% - BEST\\ 24.87\% - ER\\ \underline{25.57\% - REG}\end{tabular} & \begin{tabular}[c]{@{}c@{}}\underline{38.73\% - AVG}\\ 23.63\% - BEST\\ N/A - ER\\ N/A - REG\end{tabular}         & \begin{tabular}[c]{@{}c@{}}19.52\% - AVG\\ 20.34\% - BEST\\ \underline{23.53\% - ER}\\ 18.83\% - REG\end{tabular} \\ \hline
MOBILE       & \begin{tabular}[c]{@{}c@{}}46.62\% - AVG\\ \underline{65.11\% - BEST}\\ 58.96\% - ER\\ 57.86\% - REG\end{tabular} & \begin{tabular}[c]{@{}c@{}}45.31\% - AVG\\ 59.55\% - BEST\\ \underline{60.43\% - ER}\\ 56.49\% - REG\end{tabular} & \begin{tabular}[c]{@{}c@{}}39.44\% - AVG\\ 48.88\% - BEST\\ \underline{49.39\% - ER}\\ 46.45\% - REG\end{tabular} \\ \hline
VGG          & \begin{tabular}[c]{@{}c@{}}21.93\% - AVG\\ 61.13\% - BEST\\ 57.27\% - ER\\ \underline{62.00\% - REG}\end{tabular} & \begin{tabular}[c]{@{}c@{}}34.28\% - AVG\\ 57.48\% - BEST\\ \underline{64.25\% - ER}\\ 58.76\% - REG\end{tabular} & \begin{tabular}[c]{@{}c@{}}24.61\% - AVG\\ 13.59\% - BEST\\ \underline{24.79\% - ER}\\ 14.95\% - REG\end{tabular} \\ \hline
SURF         & \underline{23.29\%}                                                                                               & \underline{35.08\%}                                                                                               & \underline{39.62\%}                                                                                               \\ \hline
SURF\_PHASH  & \begin{tabular}[c]{@{}c@{}}21.49\% - AVG\\ 21.96\% - BEST\\ 25.68\% - ER\\ \underline{26.65\% - REG}\end{tabular} & \begin{tabular}[c]{@{}c@{}}29.22\% - AVG\\ 32.09\% - BEST\\ \underline{36.23\% - ER}\\ 33.26\% - REG\end{tabular} & \begin{tabular}[c]{@{}c@{}}20.49\% - AVG\\ 23.41\% - BEST\\ \underline{23.74\% - ER}\\ 08.77\% - REG\end{tabular} \\ \hline
SURF\_MOBILE & \begin{tabular}[c]{@{}c@{}}40.28\% - AVG\\ \underline{58.88\% - BEST}\\ 55.63\% - ER\\ 56.22\% - REG\end{tabular} & \begin{tabular}[c]{@{}c@{}}63.83\% - AVG\\ 64.32\% - BEST\\ \underline{64.96\% - ER}\\ 64.67\% - REG\end{tabular} & \begin{tabular}[c]{@{}c@{}}21.26\% - AVG\\ 22.79\% - BEST\\ 17.77\% - ER\\ \underline{26.94\% - REG}\end{tabular} \\ \hline
SURF\_VGG    & \begin{tabular}[c]{@{}c@{}}37.94\% - AVG\\ \underline{54.01\% - BEST}\\ 45.87\% - ER\\ 53.90\% - REG\end{tabular} & \begin{tabular}[c]{@{}c@{}}44.01\% - AVG\\ 55.12\% - BEST\\ 49.62\% - ER\\ \underline{57.35\% - REG}\end{tabular} & \begin{tabular}[c]{@{}c@{}}22.92\% - AVG\\ \underline{29.30\% - BEST}\\ 28.53\% - ER\\ 23.49\% - REG\end{tabular} \\ \hline
\end{tabular}
\end{table}

\begin{table}[h]
\centering
\begin{tabular}{|c|c|c|c|}
\hline
Indonesia    & Louvain                                                                                               & Markov                                                                                                & Spectral                                                                                              \\ \hline
PHASH        & \begin{tabular}[c]{@{}c@{}}\underline{32.53\% - AVG}\\ 17.90\% - BEST\\ 32.02\% - ER\\ 30.12\% - REG\end{tabular} & \begin{tabular}[c]{@{}c@{}}N/A - AVG\\ N/A - BEST\\ N/A - ER\\ N/A - REG\end{tabular}                 & \begin{tabular}[c]{@{}c@{}}31.07\% - AVG\\ \underline{31.81\% - BEST}\\ 31.61\% - ER\\ 28.45\% - REG\end{tabular} \\ \hline
MOBILE       & \begin{tabular}[c]{@{}c@{}}46.04\% - AVG\\ 58.43\% - BEST\\ 60.06\% - ER\\ \underline{65.11\% - REG}\end{tabular} & \begin{tabular}[c]{@{}c@{}}19.30\% - AVG\\ \underline{64.71\% - BEST}\\ 53.42\% - ER\\ 46.85\% - REG\end{tabular} & \begin{tabular}[c]{@{}c@{}}22.55\% - AVG\\ 32.10\% - BEST\\ 32.39\% - ER\\ \underline{35.55\% - REG}\end{tabular} \\ \hline
VGG          & \begin{tabular}[c]{@{}c@{}}34.73\% - AVG\\ 64.61\% - BEST\\ \underline{77.05\% - ER}\\ 66.92\% - REG\end{tabular} & \begin{tabular}[c]{@{}c@{}}23.72\% - AVG\\ 55.69\% - BEST\\ 52.95\% - ER\\ \underline{73.46\% - REG}\end{tabular} & \begin{tabular}[c]{@{}c@{}}30.31\% - AVG\\ 50.03\% - BEST\\ 49.44\% - ER\\ \underline{50.07\% - REG}\end{tabular} \\ \hline
SURF         & \begin{tabular}[c]{@{}c@{}}42.67\% - AVG\\ 42.36\% - BEST\\ \underline{46.08\% - ER}\\ 41.71\% - REG\end{tabular} & \begin{tabular}[c]{@{}c@{}}\underline{60.94\% - AVG}\\ 58.58\% - BEST\\ 54.73\% - ER\\ 18.18\% - REG\end{tabular} & \begin{tabular}[c]{@{}c@{}}40.70\% - AVG\\ 39.38\% - BEST\\ \underline{44.66\% - ER}\\ 39.08\% - REG\end{tabular} \\ \hline
SURF\_PHASH  & \begin{tabular}[c]{@{}c@{}}36.81\% - AVG\\ \underline{71.95\% - BEST}\\ 58.89\% - ER\\ 67.48\% - REG\end{tabular} & \begin{tabular}[c]{@{}c@{}}45.93\% - AVG\\ 81.91\% - BEST\\ \underline{86.19\% - ER}\\ 82.02\% - REG\end{tabular} & \begin{tabular}[c]{@{}c@{}}32.39\% - AVG\\ \underline{44.16\% - BEST}\\ 03.99\% - ER\\ 21.26\% - REG\end{tabular} \\ \hline
SURF\_MOBILE & \begin{tabular}[c]{@{}c@{}}25.96\% - AVG\\ 66.99\% - BEST\\ 62.78\% - ER\\ \underline{67.01\% - REG}\end{tabular} & \begin{tabular}[c]{@{}c@{}}48.19\% - AVG\\ \underline{93.81\% - BEST}\\ 88.02\% - ER\\ 86.05\% - REG\end{tabular} & \begin{tabular}[c]{@{}c@{}}\underline{45.76\% - AVG}\\ 32.26\% - BEST\\ 44.13\% - ER\\ 18.98\% - REG\end{tabular} \\ \hline
SURF\_VGG    & \begin{tabular}[c]{@{}c@{}}21.19\% - AVG\\ 38.49\% - BEST\\ 38.75\% - ER\\ \underline{43.92\% - REG}\end{tabular} & \begin{tabular}[c]{@{}c@{}}45.94\% - AVG\\ 56.61\% - BEST\\ 57.95\% - ER\\ \underline{77.68\% - REG}\end{tabular} & \begin{tabular}[c]{@{}c@{}}31.28\% - AVG\\ 21.23\% - BEST\\ 25.36\% - ER\\ \underline{31.76\% - REG}\end{tabular} \\ \hline
\end{tabular}
\end{table}

\begin{table}[h]
\centering
\begin{tabular}{|c|c|c|c|}
\hline
Ukraine      & Louvain                                                                                               & Markov                                                                                                & Spectral                                                                                              \\ \hline
PHASH        & \begin{tabular}[c]{@{}c@{}}16.98\% - AVG\\ 23.47\% - BEST\\ 23.60\% - ER\\ \underline{25.19\% - REG}\end{tabular} & \begin{tabular}[c]{@{}c@{}}N/A - AVG\\ N/A - BEST\\ N/A - ER\\ N/A - REG\end{tabular}                 & \begin{tabular}[c]{@{}c@{}}18.73\% - AVG\\ 19.36\% - BEST\\ 21.93\% - ER\\ \underline{23.22\% - REG}\end{tabular} \\ \hline
MOBILE       & \begin{tabular}[c]{@{}c@{}}49.76\% - AVG\\ 66.43\% - BEST\\ 61.99\% - ER\\ \underline{73.04\% - REG}\end{tabular} & \begin{tabular}[c]{@{}c@{}}32.43\% - AVG\\ \underline{67.68\% - BEST}\\ 65.37\% - ER\\ 56.56\% - REG\end{tabular} & \begin{tabular}[c]{@{}c@{}}23.66\% - AVG\\ \underline{55.47\% - BEST}\\ 47.97\% - ER\\ 41.21\% - REG\end{tabular} \\ \hline
VGG          & \begin{tabular}[c]{@{}c@{}}50.10\% - AVG\\ \underline{56.08\% - BEST}\\ 47.45\% - ER\\ 48.96\% - REG\end{tabular} & \begin{tabular}[c]{@{}c@{}}25.91\% - AVG\\ \underline{63.24\% - BEST}\\ 51.01\% - ER\\ 49.96\% - REG\end{tabular} & \begin{tabular}[c]{@{}c@{}}24.68\% - AVG\\ 39.51\% - BEST\\ \underline{45.26\% - ER}\\ 35.25\% - REG\end{tabular} \\ \hline
SURF         & \begin{tabular}[c]{@{}c@{}}32.86\% - AVG\\ 32.86\% - BEST\\ 32.29\% - ER\\ \underline{38.61\% - REG}\end{tabular} & \begin{tabular}[c]{@{}c@{}}19.51\% - AVG\\ \underline{57.36\% - BEST}\\ 49.56\% - ER\\ 47.15\% - REG\end{tabular} & \begin{tabular}[c]{@{}c@{}}\underline{33.68\% - AVG}\\ 27.11\% - BEST\\ 20.64\% - ER\\ 25.35\% - REG\end{tabular} \\ \hline
SURF\_PHASH  & \begin{tabular}[c]{@{}c@{}}14.39\% - AVG\\ \underline{38.35\% - BEST}\\ 29.97\% - ER\\ 37.83\% - REG\end{tabular} & \begin{tabular}[c]{@{}c@{}}39.89\% - AVG\\ \underline{65.60\% - BEST}\\ 62.26\% - ER\\ 58.62\% - REG\end{tabular} & \begin{tabular}[c]{@{}c@{}}\underline{21.78\% - AVG}\\ 20.28\% - BEST\\ 05.43\% - ER\\ 10.09\% - REG\end{tabular} \\ \hline
SURF\_MOBILE & \begin{tabular}[c]{@{}c@{}}17.35\% - AVG\\ \underline{71.55\% - BEST}\\ 54.51\% - ER\\ 66.05\% - REG\end{tabular} & \begin{tabular}[c]{@{}c@{}}44.18\% - AVG\\ 13.61\% - BEST\\ \underline{79.91\% - ER}\\ 75.77\% - REG\end{tabular} & \begin{tabular}[c]{@{}c@{}}28.89\% - AVG\\ 37.12\% - BEST\\ \underline{59.02\% - ER}\\ 14.52\% - REG\end{tabular} \\ \hline
SURF\_VGG    & \begin{tabular}[c]{@{}c@{}}31.91\% - AVG\\ 56.52\% - BEST\\ 46.76\% - ER\\ \underline{69.95\% - REG}\end{tabular} & \begin{tabular}[c]{@{}c@{}}45.82\% - AVG\\ \underline{77.42\% - BEST}\\ 75.35\% - ER\\ 75.15\% - REG\end{tabular} & \begin{tabular}[c]{@{}c@{}}30.74\% - AVG\\ 06.81\% - BEST\\ 19.88\% - ER\\ \underline{31.81\% - REG}\end{tabular} \\ \hline
\end{tabular}
\end{table}

\clearpage

\section{Cluster Structures.}\label{app:clust_structs}

\subsection{Cluster Statistics.}

\begin{table}[h]
\centering
\begin{tabular}{|c|c|c|c|}
\hline
Reddit       & Louvain                                                                               & Markov                                                                                        & Spectral                                                                              \\ \hline
PHASH        & \begin{tabular}[c]{@{}c@{}}244 - AVG\\ 257 - BEST\\ 244 - ER\\ \underline{256 - REG}\end{tabular} & \begin{tabular}[c]{@{}c@{}}\underline{10586 - AVG}\\ 10586 - BEST\\ 10586 - ER\\ 10586 - REG\end{tabular} & \begin{tabular}[c]{@{}c@{}}150 - AVG\\ 150 - BEST\\ \underline{150 - ER}\\ 150 - REG\end{tabular} \\ \hline
MOBILE       & \begin{tabular}[c]{@{}c@{}}164 - AVG\\ \underline{257 - BEST}\\ 238 - ER\\ 128 - REG\end{tabular} & \begin{tabular}[c]{@{}c@{}}355 - AVG\\ 394 - BEST\\ \underline{393 - ER}\\ 161 - REG\end{tabular}         & \begin{tabular}[c]{@{}c@{}}150 - AVG\\ 150 - BEST\\ \underline{150 - ER}\\ 150 - REG\end{tabular} \\ \hline
VGG          & \begin{tabular}[c]{@{}c@{}}127 - AVG\\ 255 - BEST\\ 236 - ER\\ \underline{256 - REG}\end{tabular} & \begin{tabular}[c]{@{}c@{}}809 - AVG\\ 425 - BEST\\ \underline{424 - ER}\\ 425 - REG\end{tabular}         & \begin{tabular}[c]{@{}c@{}}150 - AVG\\ 150 - BEST\\ \underline{150 - ER}\\ 150 - REG\end{tabular} \\ \hline
SURF         & \begin{tabular}[c]{@{}c@{}}28 - AVG\\ 28 - BEST\\ 28 - ER\\ 28 - REG\end{tabular}     & \begin{tabular}[c]{@{}c@{}}760 - AVG\\ 760 - BEST\\ 760 - ER\\ 760 - REG\end{tabular}         & \begin{tabular}[c]{@{}c@{}}150 - AVG\\ 150 - BEST\\ 150 - ER\\ 150 - REG\end{tabular} \\ \hline
SURF\_PHASH  & \begin{tabular}[c]{@{}c@{}}158 - AVG\\ 537 - BEST\\ 408 - ER\\ \underline{535 - REG}\end{tabular} & \begin{tabular}[c]{@{}c@{}}4827 - AVG\\ 4263 - BEST\\ \underline{4261 - ER}\\ 4260 - REG\end{tabular}     & \begin{tabular}[c]{@{}c@{}}150 - AVG\\ 150 - BEST\\ \underline{150 - ER}\\ 150 - REG\end{tabular} \\ \hline
SURF\_MOBILE & \begin{tabular}[c]{@{}c@{}}203 - AVG\\ \underline{397 - BEST}\\ 326 - ER\\ 396 - REG\end{tabular} & \begin{tabular}[c]{@{}c@{}}5059 - AVG\\ 4705 - BEST\\ \underline{4707 - ER}\\ 4704 - REG\end{tabular}     & \begin{tabular}[c]{@{}c@{}}150 - AVG\\ 150 - BEST\\ 150 - ER\\ \underline{150 - REG}\end{tabular} \\ \hline
SURF\_VGG    & \begin{tabular}[c]{@{}c@{}}173 - AVG\\ \underline{394 - BEST}\\ 319 - ER\\ 391 - REG\end{tabular} & \begin{tabular}[c]{@{}c@{}}4388 - AVG\\ 3905 - BEST\\ 3904 - ER\\ \underline{3905 - REG}\end{tabular}     & \begin{tabular}[c]{@{}c@{}}150 - AVG\\ \underline{150 - BEST}\\ 150 - ER\\ 150 - REG\end{tabular} \\ \hline
\end{tabular}
\caption{\label{tab:reddit_cluster_nums}The number of clusters produced from each of the 52 combinations on the Reddit data set. The number that correlates with the combination that achieved the top accuracy score on the Imposter-Host task is underlined.}
\end{table}

\begin{table}[h]
\centering
\begin{tabular}{|c|c|c|c|}
\hline
Indonesia    & Louvain                                                                                  & Markov                                                                                        & Spectral                                                                              \\ \hline
PHASH        & \begin{tabular}[c]{@{}c@{}}\underline{256 - AVG}\\ 257 - BEST\\ 256 - ER\\ 256 - REG\end{tabular}    & \begin{tabular}[c]{@{}c@{}}44612 - AVG\\ 44612 - BEST\\ 44612 - ER\\ 44612 - REG\end{tabular} & \begin{tabular}[c]{@{}c@{}}150 - AVG\\ \underline{150 - BEST}\\ 144 - ER\\ 147 - REG\end{tabular} \\ \hline
MOBILE       & \begin{tabular}[c]{@{}c@{}}159 - AVG\\ 256 - BEST\\ 254 - ER\\ \underline{256 - REG}\end{tabular}    & \begin{tabular}[c]{@{}c@{}}2264 - AVG\\ \underline{3187 - BEST}\\ 3187 - ER\\ 3186 - REG\end{tabular}     & \begin{tabular}[c]{@{}c@{}}150 - AVG\\ 150 - BEST\\ 150 - ER\\ \underline{150 - REG}\end{tabular} \\ \hline
VGG          & \begin{tabular}[c]{@{}c@{}}154 - AVG\\ 257 - BEST\\ \underline{254 - ER}\\ 256 - REG\end{tabular}    & \begin{tabular}[c]{@{}c@{}}3157 - AVG\\ 1609 - BEST\\ 1590 - ER\\ \underline{1607 - REG}\end{tabular}     & \begin{tabular}[c]{@{}c@{}}150 - AVG\\ 148 - BEST\\ 144 - ER\\ \underline{149 - REG}\end{tabular} \\ \hline
SURF         & \begin{tabular}[c]{@{}c@{}}69 - AVG\\ 72 - BEST\\ \underline{68 - ER}\\ 73 - REG\end{tabular}        & \begin{tabular}[c]{@{}c@{}}\underline{3103 - AVG}\\ 3136 - BEST\\ 3103 - ER\\ 3103 - REG\end{tabular}     & \begin{tabular}[c]{@{}c@{}}150 - AVG\\ 150 - BEST\\ \underline{150 - ER}\\ 150 - REG\end{tabular} \\ \hline
SURF\_PHASH  & \begin{tabular}[c]{@{}c@{}}197 - AVG\\ \underline{1456 - BEST}\\ 846 - ER\\ 1609 - REG\end{tabular}  & \begin{tabular}[c]{@{}c@{}}16197 - AVG\\ 13659 - BEST\\ \underline{13620 - ER}\\ 13648 - REG\end{tabular} & \begin{tabular}[c]{@{}c@{}}150 - AVG\\ \underline{146 - BEST}\\ 147 - ER\\ 150 - REG\end{tabular} \\ \hline
SURF\_MOBILE & \begin{tabular}[c]{@{}c@{}}183 - AVG\\ 1597 - BEST\\ 846 - ER\\ \underline{1609 - REG}\end{tabular}  & \begin{tabular}[c]{@{}c@{}}17531 - AVG\\ \underline{14670 - BEST}\\ 14639 - ER\\ 14668 - REG\end{tabular} & \begin{tabular}[c]{@{}c@{}}150 - AVG\\ \underline{149 - BEST}\\ 146 - ER\\ 150 - REG\end{tabular} \\ \hline
SURF\_VGG    & \begin{tabular}[c]{@{}c@{}}154 - AVG\\ 2000 - BEST\\ 1150 - ER\\ \underline{2008 - REG}\end{tabular} & \begin{tabular}[c]{@{}c@{}}15280 - AVG\\ 11712 - BEST\\ 11687 - ER\\ \underline{11703 - REG}\end{tabular} & \begin{tabular}[c]{@{}c@{}}150 - AVG\\ 149 - BEST\\ 150 - ER\\ \underline{150 - REG}\end{tabular} \\ \hline
\end{tabular}
\caption{\label{tab:indo_cluster_nums}The number of clusters produced from each of the 52 combinations on the Indonesia data set.}
\end{table}

\begin{table}[h]
\centering
\begin{tabular}{|c|c|c|c|}
\hline
Ukraine      & Louvain                                                                                & Markov                                                                                        & Spectral                                                                              \\ \hline
PHASH        & \begin{tabular}[c]{@{}c@{}}252 - AVG\\ 257 - BEST\\ 252 - ER\\ \underline{256 - REG}\end{tabular}  & \begin{tabular}[c]{@{}c@{}}16433 - AVG\\ 16433 - BEST\\ 16433 - ER\\ 16433 - REG\end{tabular} & \begin{tabular}[c]{@{}c@{}}150 - AVG\\ 150 - BEST\\ 148 - ER\\ \underline{147 - REG}\end{tabular} \\ \hline
MOBILE       & \begin{tabular}[c]{@{}c@{}}162 - AVG\\ 256 - BEST\\ 257 - ER\\ \underline{256 - REG}\end{tabular}  & \begin{tabular}[c]{@{}c@{}}416 - AVG\\ \underline{511 - BEST}\\ 501 - ER\\ 510 - REG\end{tabular}         & \begin{tabular}[c]{@{}c@{}}150 - AVG\\ \underline{149 - BEST}\\ 148 - ER\\ 150 - REG\end{tabular} \\ \hline
VGG          & \begin{tabular}[c]{@{}c@{}}138 - AVG\\ \underline{252 - BEST}\\ 238 - ER\\ 256 - REG\end{tabular}  & \begin{tabular}[c]{@{}c@{}}1068 - AVG\\ \underline{437 - BEST}\\ 437 - ER\\ 436 - REG\end{tabular}        & \begin{tabular}[c]{@{}c@{}}150 - AVG\\ 150 - BEST\\ \underline{150 - ER}\\ 150 - REG\end{tabular} \\ \hline
SURF         & \begin{tabular}[c]{@{}c@{}}17 - AVG\\ 18 - BEST\\ 17 - ER\\ \underline{21 - REG}\end{tabular}      & \begin{tabular}[c]{@{}c@{}}2169 - AVG\\ \underline{2169 - BEST}\\ 2169 - ER\\ 2169 - REG\end{tabular}     & \begin{tabular}[c]{@{}c@{}}\underline{149 - AVG}\\ 148 - BEST\\ 150 - ER\\ 148 - REG\end{tabular} \\ \hline
SURF\_PHASH  & \begin{tabular}[c]{@{}c@{}}94 - AVG\\ \underline{1203 - BEST}\\ 694 - ER\\ 1202 - REG\end{tabular} & \begin{tabular}[c]{@{}c@{}}8398 - AVG\\ \underline{5453 - BEST}\\ 5449 - ER\\ 5451 - REG\end{tabular}     & \begin{tabular}[c]{@{}c@{}}\underline{150 - AVG}\\ 148 - BEST\\ 148 - ER\\ 150 - REG\end{tabular} \\ \hline
SURF\_MOBILE & \begin{tabular}[c]{@{}c@{}}97 - AVG\\ \underline{1282 - BEST}\\ 658 - ER\\ 1286 - REG\end{tabular} & \begin{tabular}[c]{@{}c@{}}9084 - AVG\\ 5730 - BEST\\ \underline{5722 - ER}\\ 5727 - REG\end{tabular}     & \begin{tabular}[c]{@{}c@{}}150 - AVG\\ 149 - BEST\\ \underline{147 - ER}\\ 150 - REG\end{tabular} \\ \hline
SURF\_VGG    & \begin{tabular}[c]{@{}c@{}}98 - AVG\\ 1191 - BEST\\ 645 - ER\\ \underline{1189 - REG}\end{tabular} & \begin{tabular}[c]{@{}c@{}}8351 - AVG\\ \underline{5523 - BEST}\\ 5511 - ER\\ 5520 - REG\end{tabular}     & \begin{tabular}[c]{@{}c@{}}150 - AVG\\ 146 - BEST\\ 148 - ER\\ \underline{150 - REG}\end{tabular} \\ \hline
\end{tabular}
\caption{\label{tab:ukr_cluster_nums}The number of clusters produced from each of the 52 combinations on the Ukraine data set.}
\end{table}

\clearpage

\begin{figure}[h]
    \centering
    \includegraphics[width=\linewidth]{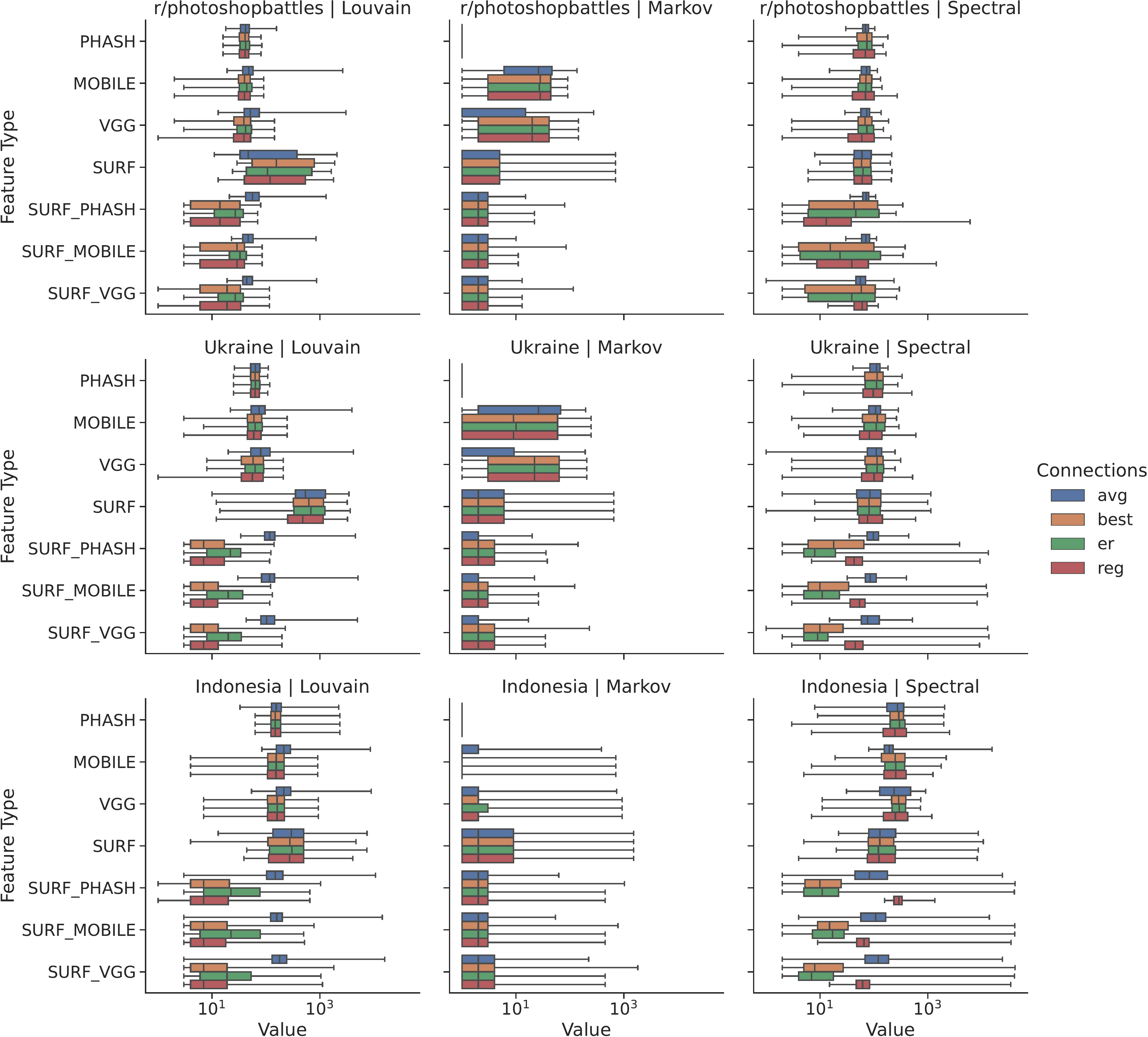}
    \caption{Box plots of the distribution of cluster sizes for each data set and each combination of feature type, clustering algorithm, and connection type. Note that the $x$-axis has a logarithmic scale.}
    \label{fig:cluster_size_boxplots}
\end{figure}

\subsection{Cluster Image Distributions.}\label{app:img_distr}

As the motif mining pipeline is intended to aid and be consumed by human observers, we believe the distribution of images amongst the clusters is of the utmost importance. Figure~\ref{fig:cluster_size_boxplots} shows box and whisker plots for all of the possible combinations. While the Markov clustering algorithm delivers the highest accuracy scores on the imposter host test, it is important to realize that the majority of the clusters are of size 1, or in other words useless to reviewers. The highest realized accuracy score was SURF\_MOBILE-BEST-MARKOV on the Indonesian data set. However, the second quartile for the image distribution was at 2 images per cluster and the third quartile is only 3 images per cluster. Out of these clusters only 63.38\% were of a size larger than 1, and only 20.59\% contained more than 3 images (I.E. valid for the Imposter-Host task). From Figure~\ref{fig:cluster_size_boxplots} we can see that this trend holds for almost all possible combinations when Markov clustering is used. It is for this reason that we recommend Louvain clustering used with the globally tagged features. In contrast to the Markov statistics, SURF\_MOBILE-BEST-LOUVAIN, on the Indonesian data set, has a second quartile at 7 images and the third quartile is 19 images. Additionally 100\% of the clusters have more than 1 image per cluster and 78.46\% have more than 3 images. Per Figure~\ref{fig:cluster_size_boxplots} this trend holds similar for all combinations and on all three data sets.

If one were to look at just Figure~\ref{fig:cluster_size_boxplots} they might come to the conclusion that Spectral clustering achieves a similar distribution to Louvain clustering and may wonder why the authors recommend Louvain clustering over Spectral clustering. It is for this reason that the whiskers are important. The maximum cluster size for SURF\_MOBILE-BEST-SPECTRAL is 40,909 images. The data set contains 44,612 images. With 40,909 images in a single cluster this means that 91.69\% of the images are essentially unsorted. We consider this case more or less useless to human reviewers, in much the same way as Markov clustering putting thousands of images into their own clusters. It is for these reasons that we believe Louvain clustering is the best of the three methods tested for motif mining.

\clearpage

\section{Graph Structures.}\label{app:graph_structs}

\begin{table}[h]
\centering
\begin{tabular}{|c|c|c|c|}
\hline
Components, Edges & Reddit       & Ukr           & Indo          \\ \hline
PHASH             & 256C, 38068E & 256C, 58537E  & 256C, 908593E \\ \hline
MOBILE            & 256C, 38523E & 256C, 58565E  & 256C, 202405E \\ \hline
VGG               & 256C, 41440E & 256C, 63952E  & 256C, 193731E \\ \hline
SURF              & 1C, 161253E  & 1C, 197858E   & 14C, 475000E  \\ \hline
SURF\_PHASH       & 412C, 24877E & 935C, 35938E  & 1085C, 209128E\\ \hline
SURF\_MOBILE      & 336C, 21728E & 1112C, 32887E & 1237C, 203859E\\ \hline
SURF\_VGG         & 324C, 18837E & 984C, 33389E  &   1372C, 213988E              \\ \hline
\end{tabular}
\caption{\label{tab:comps_edges}The number of components and edges the generated graph contained for each feature type for each data set. Of particular interest is each global feature resulting in 256 components (due to the number of FAISS centroids), SURF features producing 1, 1, and 14 components (due to their locality and diversity of query results), and the married features resulting in a relatively high number of components implying the discovery of 'sub-structures' of similar images within the already calculated FAISS centroids.}
\end{table}

\subsection{Centroid and Tag Number Experiments.}\label{app:cent_tag}

\begin{table}[h]
\centering
\begin{tabular}{|c|c|c|c|c|}
\hline
Components, Edges & 128 Centroids & 256 Centroids & 512 Centroids & 1024 Centroids \\ \hline
PHASH             & 128C, 50406E  & 256C, 38068E  & 512C, 25491E  & 1024C, 19200E  \\ \hline
MOBILE            & 128C, 59138E  & 256C, 38523E  & 512C, 26390E  & 1024C, 19937E  \\ \hline
VGG               & 128C, 68027E  & 256C, 41440E  & 512C, 27070E  & 1024C, 19652E  \\ \hline
SURF              & 1C, 158000E   & 1C, 161253E   & 1C, 159750E   & 1C, 157600E    \\ \hline
SURF\_PHASH       & 233C, 24753E  & 412C, 24877E  & 733C, 24086E  & 1257C, 22406E  \\ \hline
SURF\_MOBILE      & 205C, 22353E  & 336C, 21728E  & 599C, 21466E  & 1116C, 20432E  \\ \hline
SURF\_VGG         & 200C, 19667E  & 324C, 18837E  & 588C, 19008E  & 1083C, 18039E  \\ \hline
\end{tabular}
\caption{\label{tab:comps_edges_centroids}The number of components and edges the resulting graphs had when the index was created with 128, 256, 512, and 1024 centroids. This shows that regardless of the number of centroids chosen all the global features accomplish is exposing the pre-existing centroid space from the OPQ index.}
\end{table}

\begin{table}[h]
\centering
\begin{tabular}{|c|c|c|c|c|}
\hline
Components, Edges & 8 Length Tag & 16 Length Tag & 32 Length Tag & 64 Length Tag \\ \hline
SURF\_PHASH       & 259C, 27575E & 412C, 24877E  & N/A           & N/A           \\ \hline
SURF\_MOBILE      & 362C, 20789E & 336C, 21728E  & 633C, 18447E  & 577C, 18706E  \\ \hline
SURF\_VGG         & 340C, 18777E & 324C, 18837E  & 558C, 16040E  & 533C, 16189E  \\ \hline
\end{tabular}
\caption{\label{tab:comps_edges_tags}How the length of the global tag affects the number of components and edges in the resulting graph. The fact that PHASH features have a length of 16 was the primary driver of that length being used. One can see however that increasing the tag almost doubles the number of components between 16 and 32. If the goal is a larger number of discrete clusters this might be a worthwhile change.}
\end{table}

\clearpage

\section{Extra Figures and Data}

\subsection{Meme Clusters and Examples}

\begin{figure}[h]
    \centering
    \includegraphics[width=\linewidth]{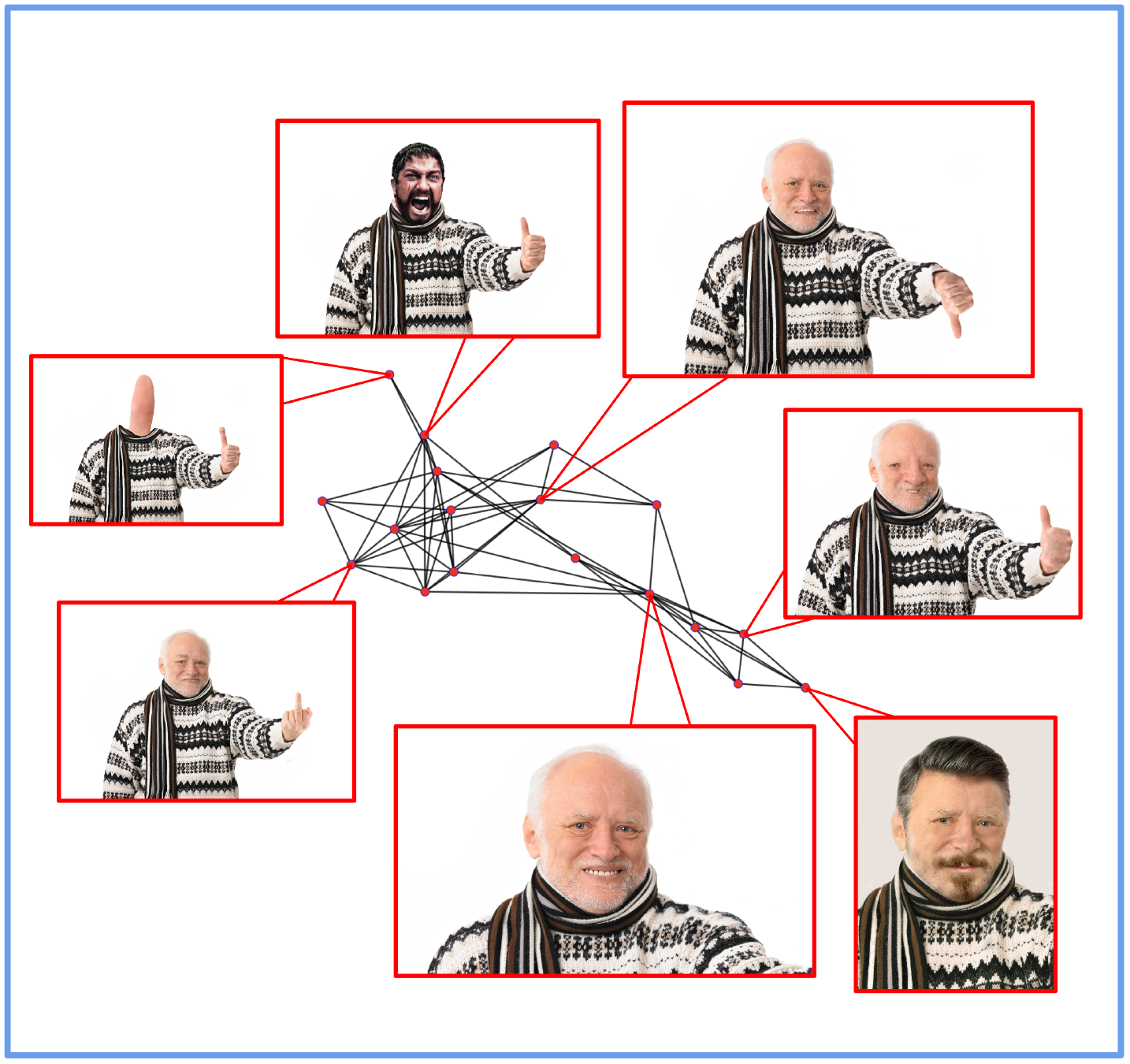}
    \caption{An example of a Reddit cluster demonstrating the kind of visual remixing done in the data set.}
    \label{fig:reddit_cluster_harold.png}
\end{figure}

\begin{figure}[h]
    \centering
    \includegraphics[width=\linewidth]{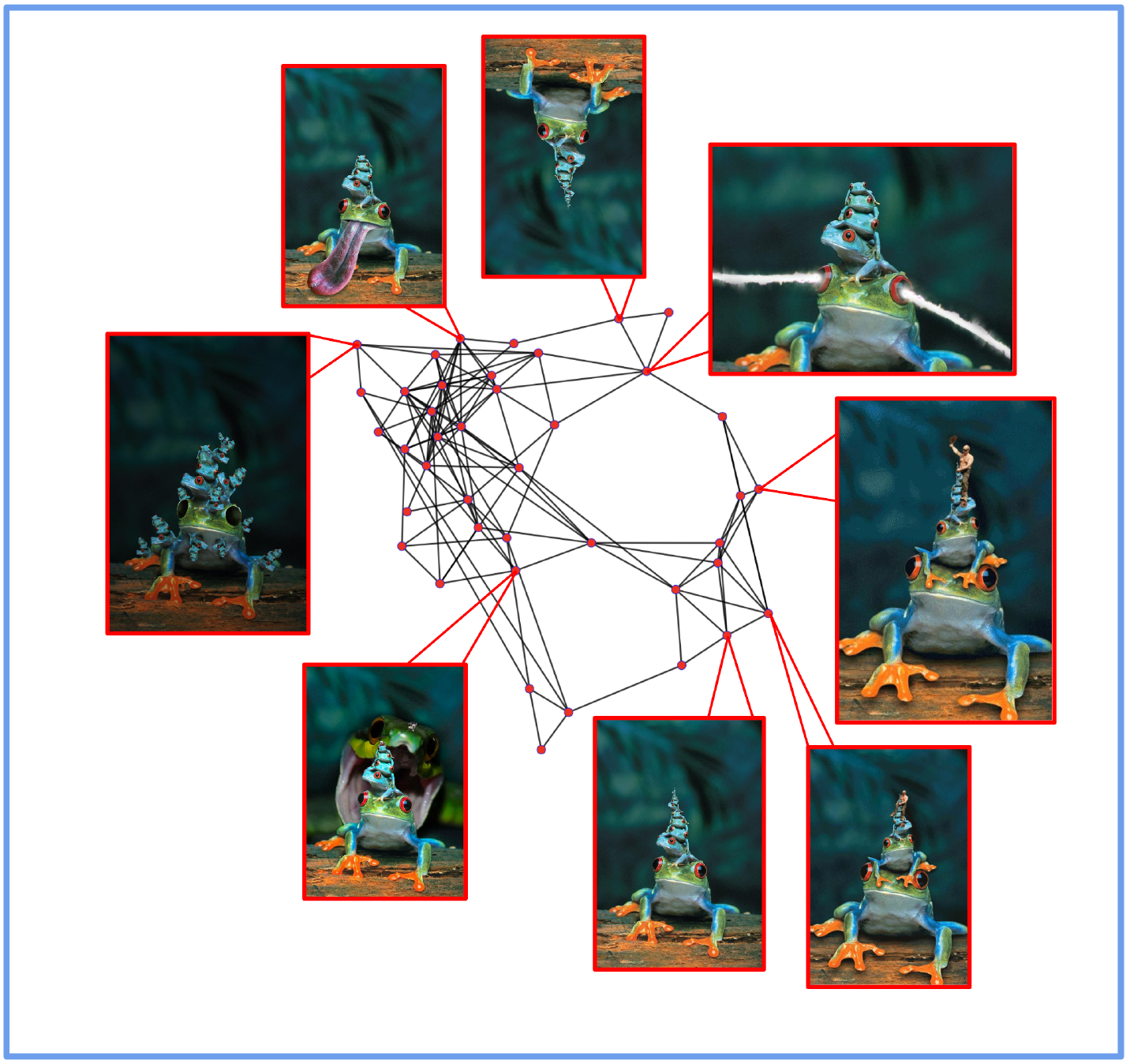}
    \caption{A cluster containing remixes of a stack of tree frogs. This cluster shows the usefulness of the global tag, as all the images look very similar globally and their matching can benefit from the composite feature type.}
    \label{fig:reddit_cluster_frogs.png}
\end{figure}

\begin{figure}[h]
    \centering
    \includegraphics[width=\linewidth]{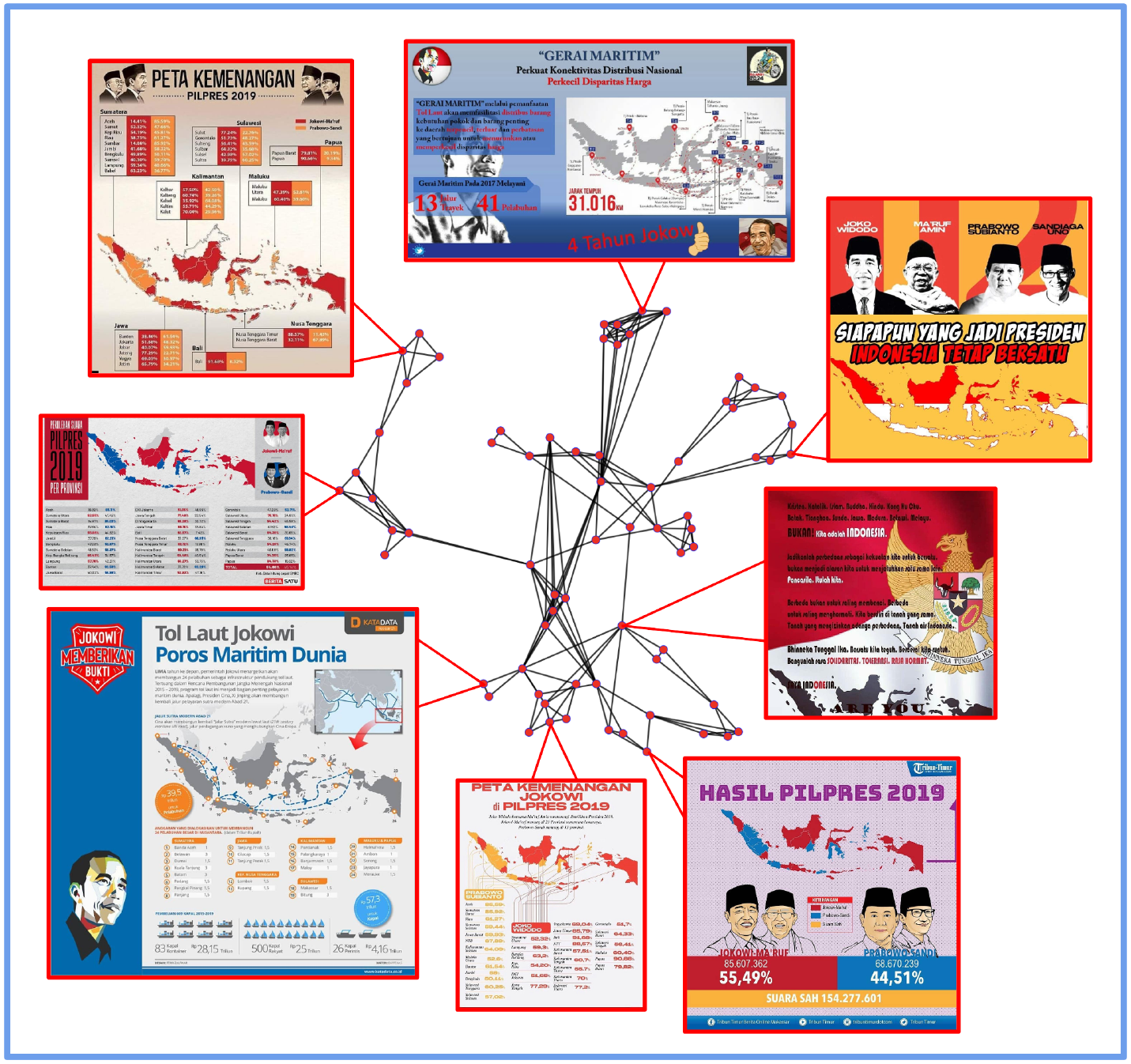}
    \caption{An example of the local features being used to create a cluster on the Indonesian data set. Each image, while globally very different, contains at least part of a map of Indonesia. The local features are able to find the shared map portions in each of the images and cluster them together.}
    \label{fig:indo_maps}
\end{figure}

\begin{figure}[h]
    \centering
    \includegraphics[width=\linewidth]{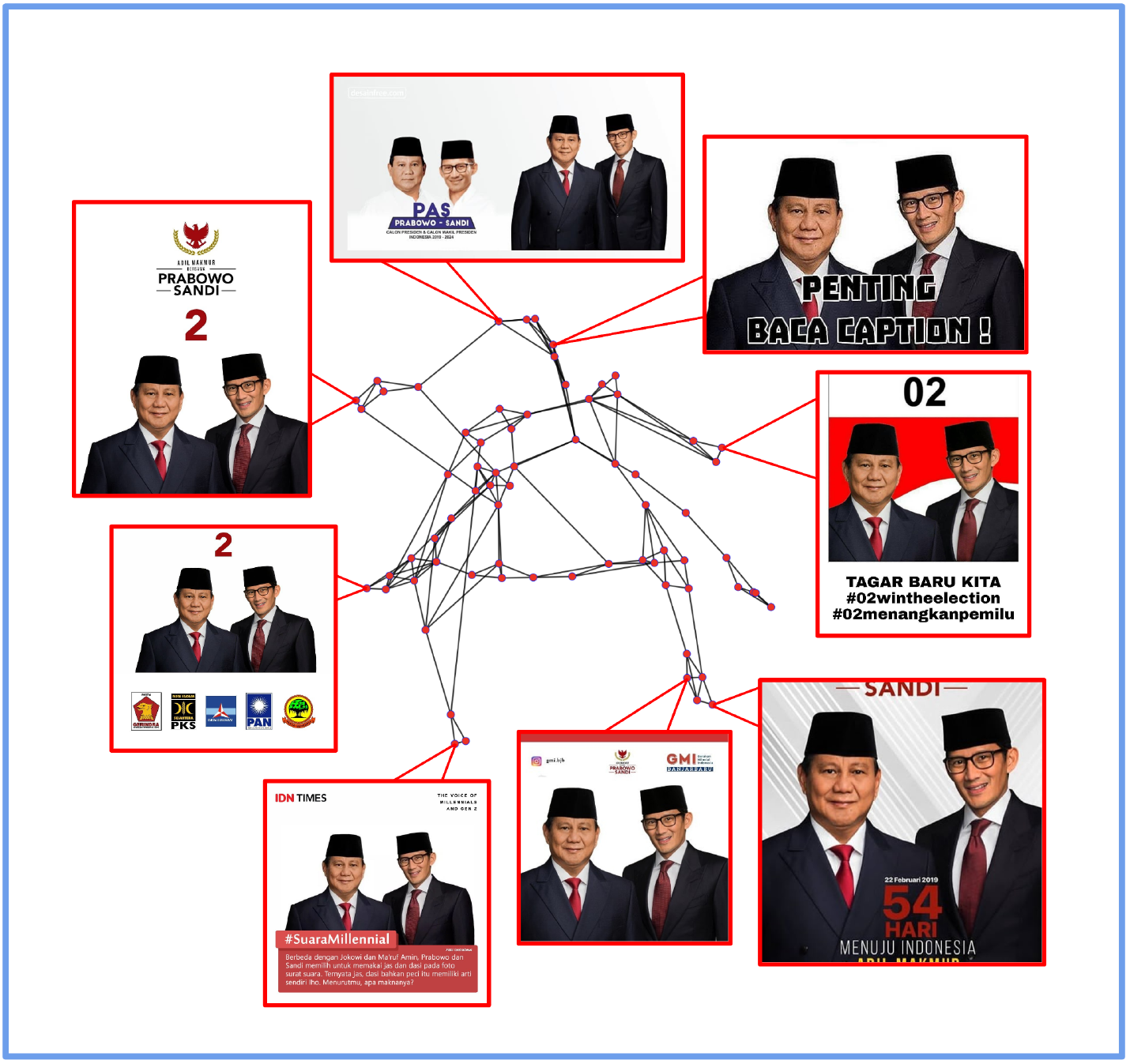}
    \caption{A collection of presidential campaign ads from the Indonesian election in 2019. The same base image is used throughout but is remixed in various contexts. This kind of campaign ad remixing was common in the data set for both candidates.}
    \label{fig:indo_prabo}
\end{figure}

\begin{figure}[h]
    \centering
    \includegraphics[width=\linewidth]{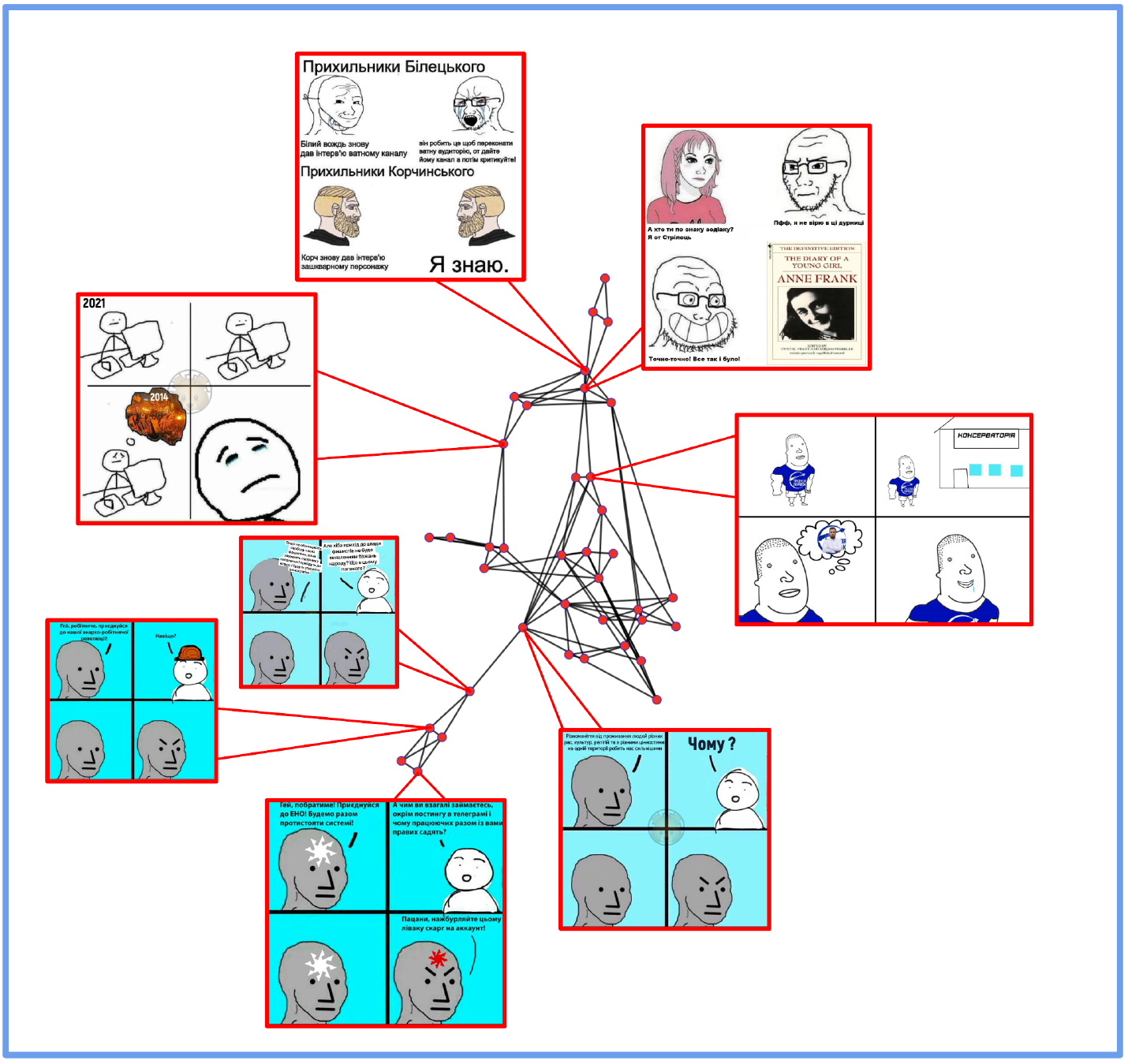}
    \caption{A cluster of Ukrainian memes. While initially appearing different all of the memes share the same four panel structure and a subgroup of them share the same genre on top of which various topics are remixed.}
    \label{fig:ukr_npcs}
\end{figure}

\begin{figure}[h]
    \centering
    \includegraphics[width=\linewidth]{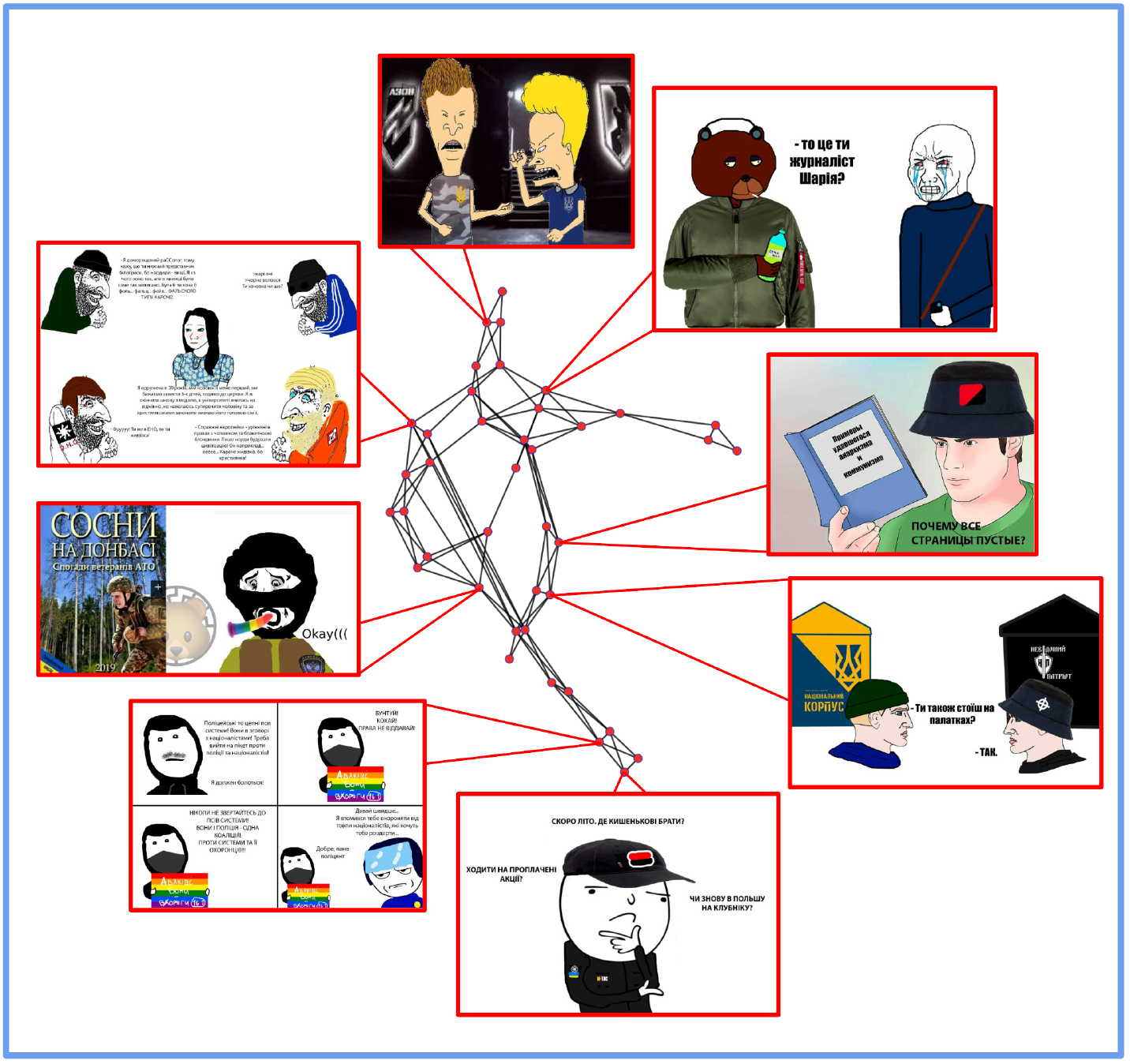}
    \caption{An example of a cluster which human observers may consider interesting but we would consider a failure from an algorithmic perspective. While human observers might be interested in exploring the online meme space on Telegram, visually the images in this cluster do not have much in common. While the implemented motif mining pipeline is good, it is far from perfect and not every cluster contains a recognizable motif from a computer vision stand-point.}
    \label{fig:ukr_memes}
\end{figure}

\clearpage

\subsection{Connection Type Plots}

\begin{figure}[h]
\centering
\includegraphics[width=0.94\textwidth]{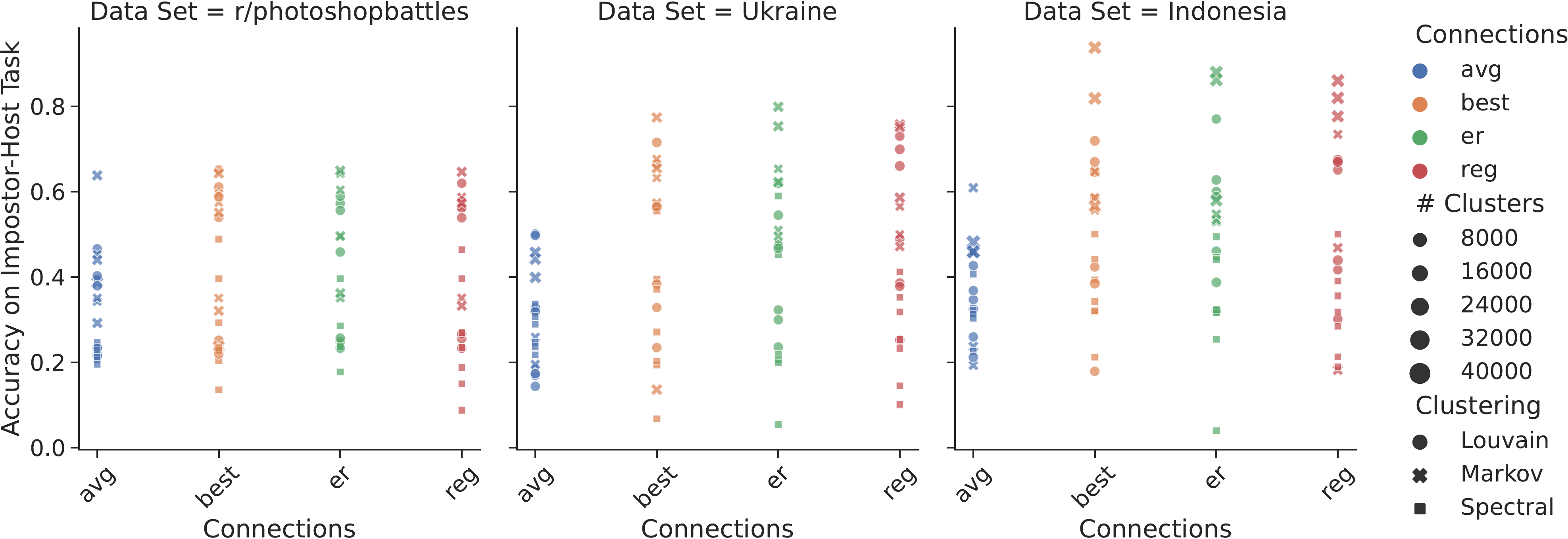} 
\caption{The accuracy scores of the Imposter-Host test across the three data sets for each connection type. Each of the three clustering methods is noted with a different shape. The size of each marker is proportional to the number of clusters.}
\label{fig:acc_comps1}
\end{figure}

\begin{figure}[h]
\centering
\includegraphics[width=0.94\textwidth]{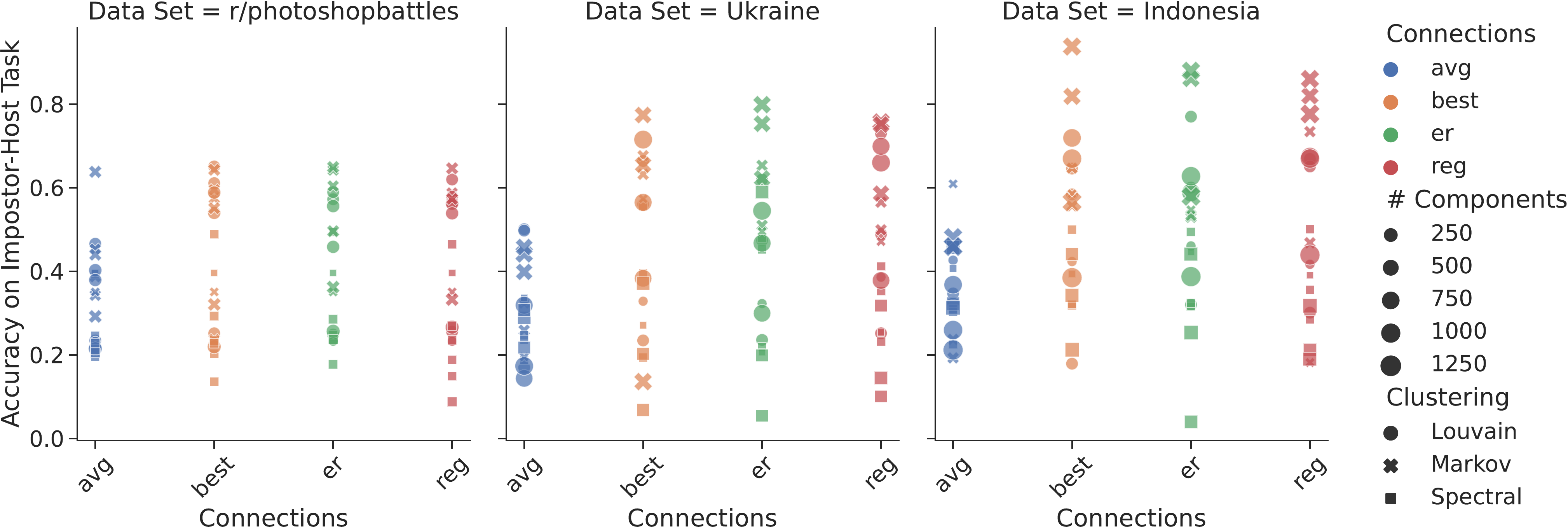} 
\caption{The accuracy scores of the Imposter-Host test across the three data sets for each connection type. Each of the three clustering methods is noted with a different shape. The size of each marker is proportional to the number of components in the graph.}
\label{fig:acc_comps2}
\end{figure}

\begin{figure}[h]
\centering
\includegraphics[width=0.94\textwidth]{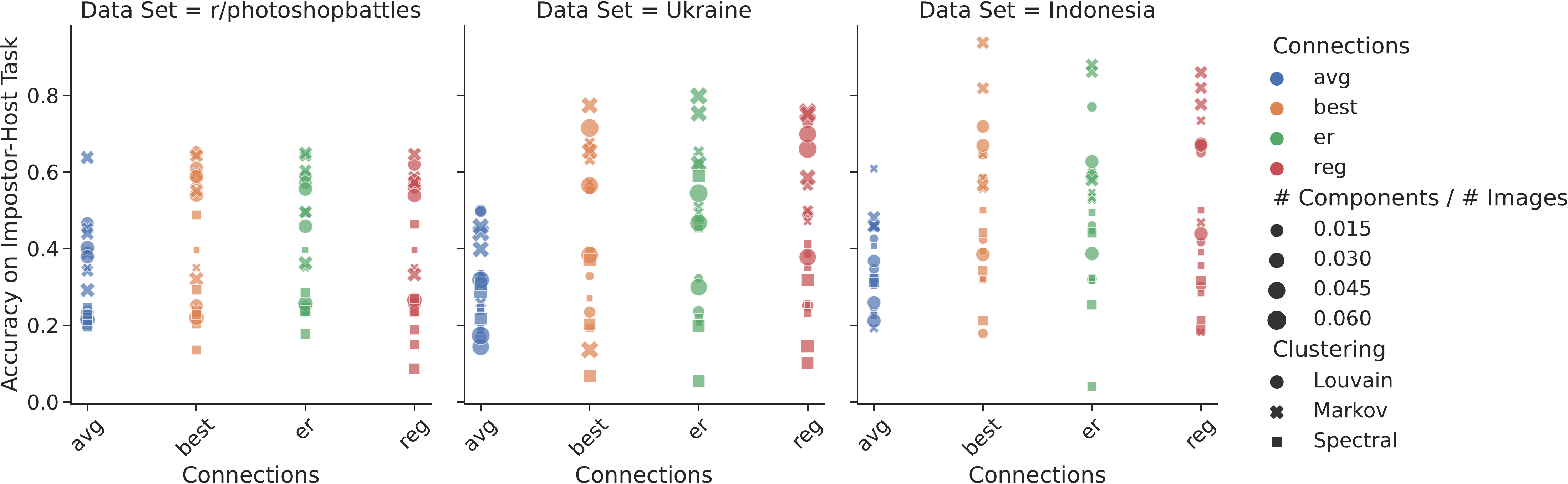} 
\caption{The accuracy scores of the Imposter-Host test across the three data sets for each connection type. Each of the three clustering methods is noted with a different shape. The size of each marker is proportional to the ratio of the number of components in the graph to the number of images in the dataset.}
\label{fig:acc_comps3}
\end{figure}

\begin{figure}[h]
\centering
\includegraphics[width=0.94\textwidth]{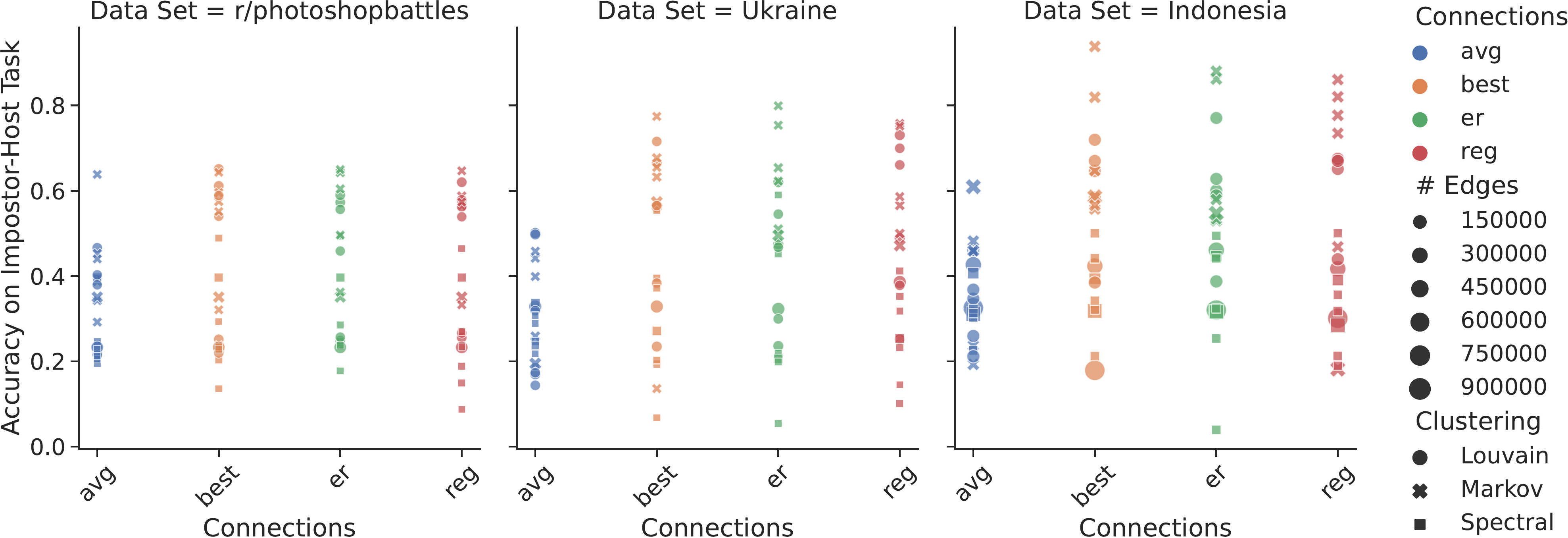} 
\caption{The accuracy scores of the Imposter-Host test across the three data sets for each connection type. Each of the three clustering methods is noted with a different shape. The size of each marker is proportional to the number of edges in the graph.}
\label{fig:acc_comps4}
\end{figure}

\clearpage

\subsection{Feature Type Plots}

\begin{figure}[h]
\centering
\includegraphics[width=0.94\textwidth]{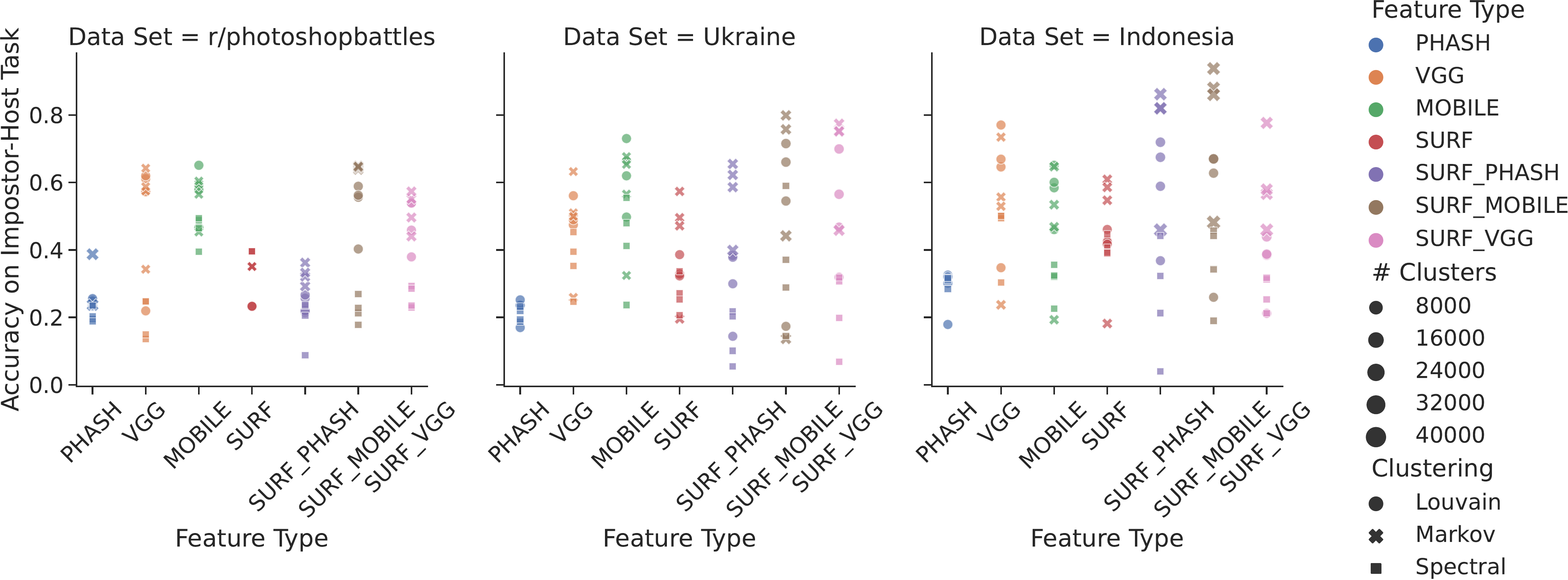} 
\caption{The accuracy scores of the Imposter-Host test across the three data sets for each feature type. Each of the three clustering methods is noted with a different shape. The size of each marker is proportional to the number of clusters.}
\label{fig:acc_comps5}
\end{figure}

\begin{figure}[h]
\centering
\includegraphics[width=0.94\textwidth]{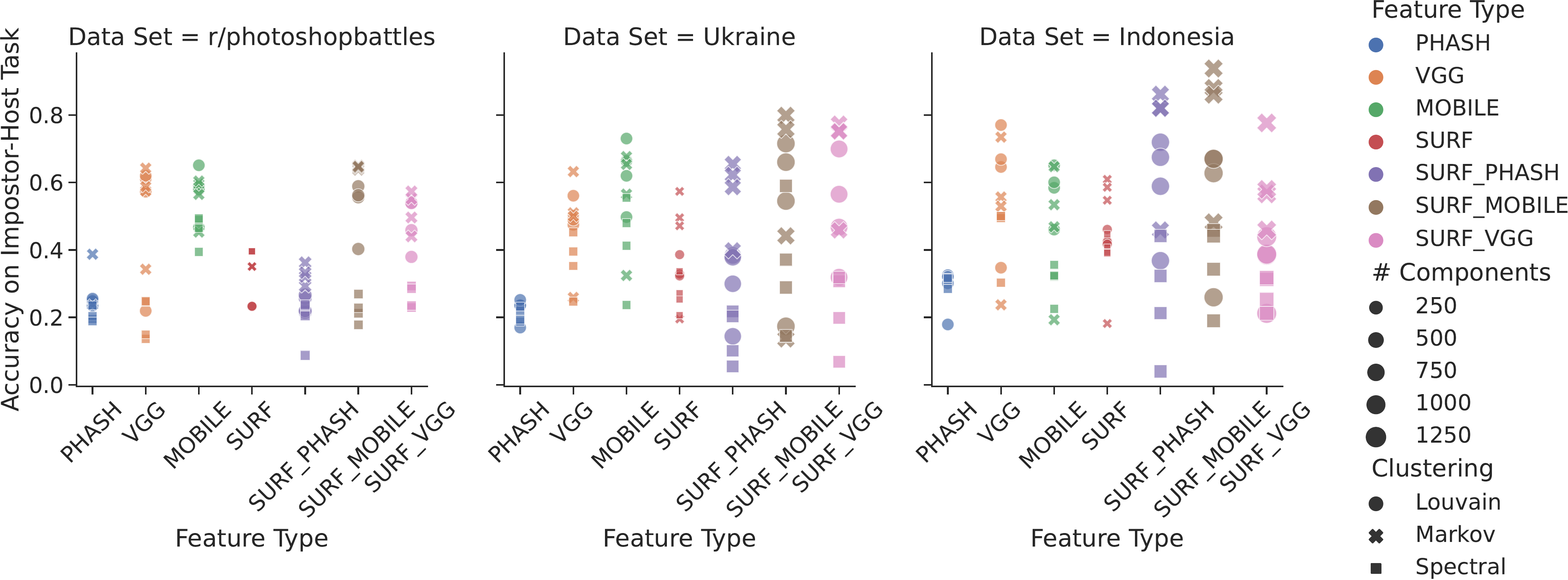} 
\caption{The accuracy scores of the Imposter-Host test across the three data sets for each feature type. Each of the three clustering methods is noted with a different shape. The size of each marker is proportional to the number of components in the graph.}
\label{fig:acc_comps6}
\end{figure}

\begin{figure}[h]
\centering
\includegraphics[width=0.94\textwidth]{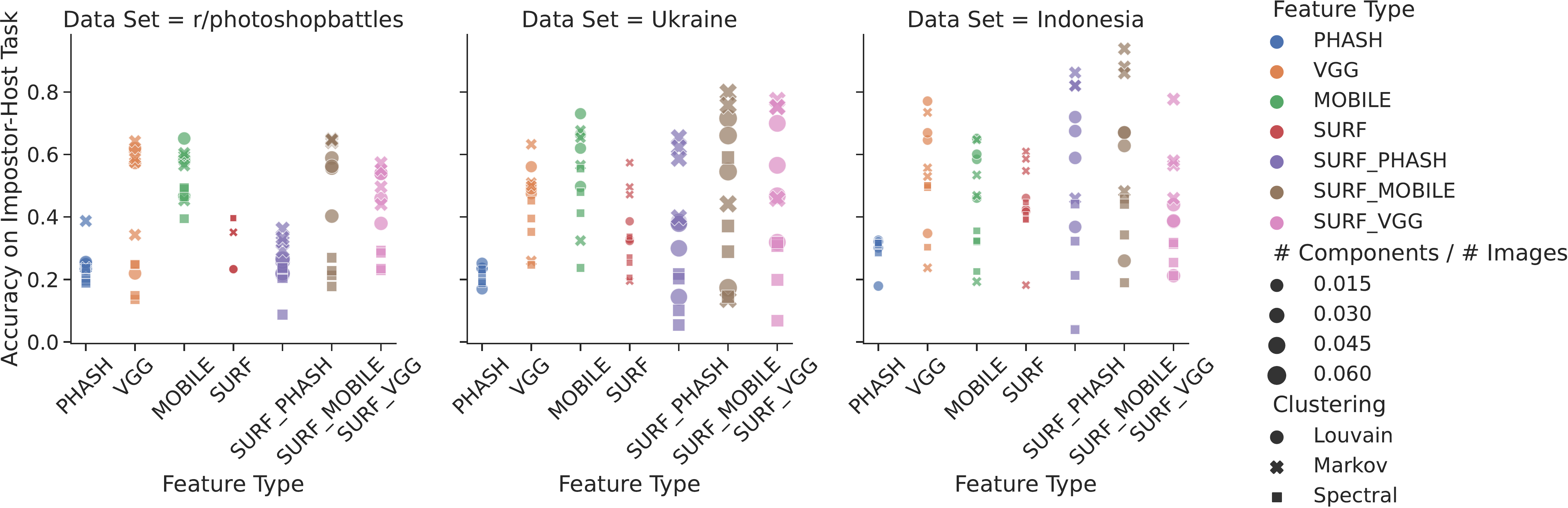} 
\caption{The accuracy scores of the Imposter-Host test across the three data sets for each feature type. Each of the three clustering methods is noted with a different shape. The size of each marker is proportional to the ratio of the number of components in the graph to the number of images in the dataset.}
\label{fig:acc_comps7}
\end{figure}

\begin{figure}[h]
\centering
\includegraphics[width=0.94\textwidth]{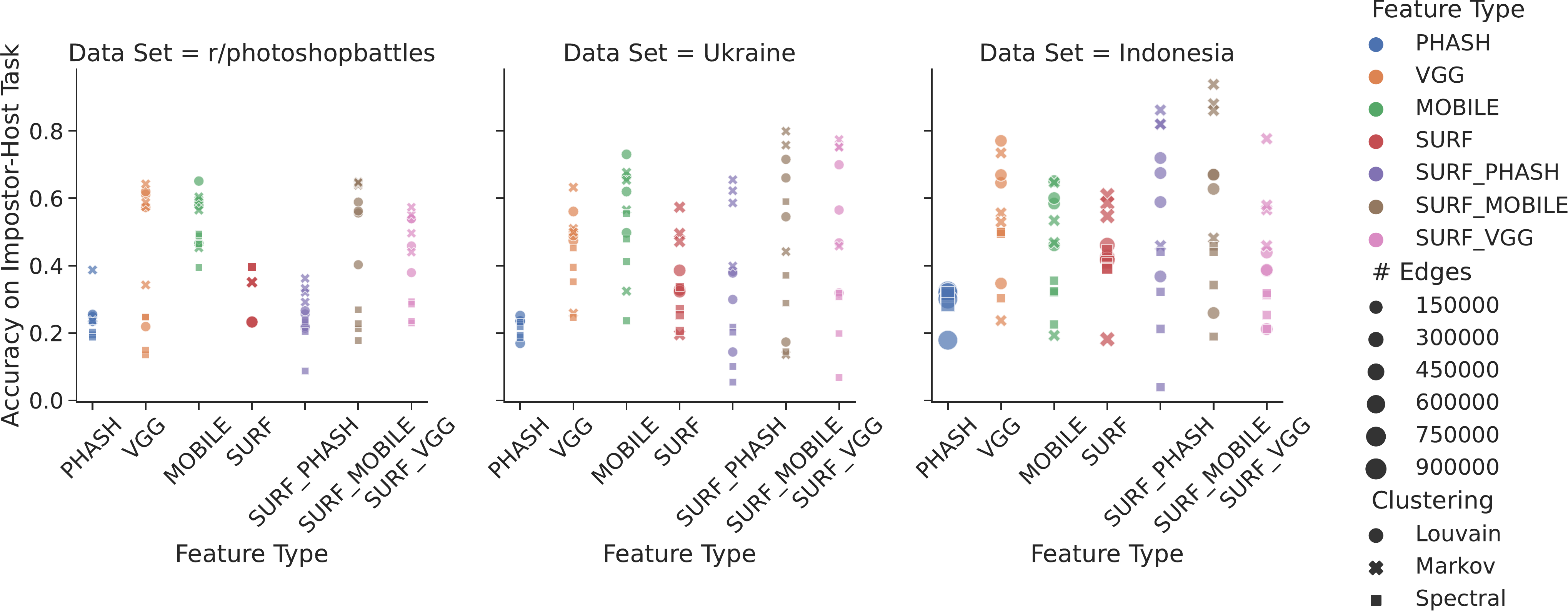} 
\caption{The accuracy scores of the Imposter-Host test across the three data sets for each feature type. Each of the three clustering methods is noted with a different shape. The size of each marker is proportional to the number of edges in the graph.}
\label{fig:acc_comps8}
\end{figure}

\clearpage

\section{Telegram Users}\label{app:users}

Medvezhatko1488, sashakots, russ\_orientalist, white\_powder2020, karpatska\_sich, NSDviz, dadzibao, olifand\_rolands, ASupersharij, BerezaJuice, dark\_k, joker\_ukr, kryuchoktv, legitimniy, notesdetective, rezident\_ua, smolii\_ukraine, \\ tayni\_deputata, thanksrinat, nationalcorps, nedotorkani, ivkolive, ze\_konets, ukrnastup, dubinskypro, ruheight, AleksandrSemchenko, botsmanua, borodatayaba, gistapa, kachuratut, poliakovanton, BeregTime, MaksymZhorin,\\ tradition\_and\_order, KlymenkoTime, sorosata, tsibulya\_ua, Ten\_NaPleten, donbasscase, lugansk\_inside, sorok40russia, ze\_landia, zv\_kyiv, moh\_zdoh, wargonzo, apleonkov, PiB88, format\_W, gribvictoria, maksnazar, sheptoon, dobkinmm, UlejUA, spletnicca, razvedinfo, rus\_demiurge, LastBP, zlobniaukr, mig41, catars\_is, ukrain1an\_news, korchynskiy, ua\_stalker, project\_solaris, liberaxy, orthodox\_news, sooproon\_bestiary, tasty\_flashbacks, fascio\_memes, intolerant\_historian, Ironvoter, mem\_lozha, knpu\_division, kekistandivision, EternalMuscovites, nt\_orthodox, intolerant\_journalist, AD\_i\_OR, nazbolukr, odindrugqoom, DeepStateUA, ukrnastup, ep867, legion\_of\_kuchma, NFafaf, History\_Q, vidardivision, avantguardia, ulpra, KARAS\_EVGEN, GrantDetector, privatnamemarnya, OstanniyCapitalist, afemina, totalopir, intermariumnc, intolerant\_warfighter, ukrmemesmineproblemes, evil\_ukraine, national\_resistance\_ua, propala\_gramota, postbased, ukrainianintolerant, korchynskiy, Ukrainianintolerantrezerv, RightLit, selo\_divisionS, mayonez\_sorosa, ubd\_ua, national\_corp\_kyiv, centuriaua

\end{document}